\pdfoutput=1
\documentclass{article}

 \usepackage[main, final]{neurips_2026}

\usepackage[utf8]{inputenc} %
\usepackage[T1]{fontenc}    %
\usepackage{url}            %
\usepackage{booktabs}
\usepackage{longtable}
\usepackage{pdflscape}       %
\usepackage{amsfonts}       %
\usepackage{nicefrac}       %
\usepackage{microtype}      %
\usepackage{enumitem}
\usepackage{float}
\usepackage{amsmath}
\usepackage{amssymb}
\usepackage{amsthm}
\usepackage{algorithm}
\usepackage{algpseudocode}
\usepackage{tikz}
\usetikzlibrary{arrows.meta}
\usepackage{amsthm}
\newtheorem{proposition}{Proposition}
\newtheorem{definition}{Definition}
\usepackage[dvipsnames]{xcolor} 
\usepackage[colorlinks=true,linkcolor=MidnightBlue,anchorcolor=black,citecolor=MidnightBlue,urlcolor=MidnightBlue, pdfborderstyle={/S/U/W 0.5}]{hyperref} 

\title{Causal Concept Explanations for Deep Neural Models}

\expandafter\ifx\csname iflatexml\endcsname\relax\expandafter\newif\csname iflatexml\endcsname\fi
\iflatexml
\author{
  Josh Rountree \\
  Computer Science and Artificial Intelligence Laboratory, MIT \\
  \texttt{rountree@mit.edu}
  \and
  Pulkit Verma \\
  Indian Institute of Technology Madras \\
  \texttt{pulkitv@cse.iitm.ac.in}
  \and
  Oswin So \\
  Laboratory for Information and Decision Systems, MIT \\
  \texttt{oswinso@mit.edu}
  \and
  Chuchu Fan \\
  Laboratory for Information and Decision Systems, MIT \\
  \texttt{chuchu@mit.edu}
  \and
  Julie A. Shah \\
  Computer Science and Artificial Intelligence Laboratory, MIT \\
  \texttt{julie\_a\_shah@csail.mit.edu}
}
\else
\author{
  Josh Rountree$^{1}$, 
  Pulkit Verma$^{2}$, 
  Oswin So$^{3}$, 
  Chuchu Fan$^{3}$,
  Julie A. Shah$^{1}$ \\[4pt]
  $^{1}$Computer Science and Artificial Intelligence Laboratory, MIT \\
  $^{2}$Indian Institute of Technology Madras \\
  $^{3}$Laboratory for Information and Decision Systems, MIT \\[4pt]
  \texttt{\{rountree, oswinso, chuchu\}@mit.edu}, \\
  \texttt{pulkitv@cse.iitm.ac.in}, \;
  \texttt{julie\_a\_shah@csail.mit.edu}
}
\fi

\begin{document}

\maketitle

\begin{abstract}

Concept-based explanations in neural models are important for the development, certification, and adoption of autonomous systems. Yet, interpretability remains correlational. Without methods to guide and quantify causal relationships between explanations, a concept explanation can be provided without being necessary or sufficient for the observed output thereby appearing correct while misleading. We introduce Causal Concept Wrapper Network (CCW-Net), the first use of interventional mediation analysis from fields such as economics, medicine, and epidemiology as a training objective to align causal effects of concept-based explanations with a model's output. We further apply mediation analysis as an evaluation tool to quantify, per concept label combination, how much of any model's output is causally attributable to each concept explanation. As a training objective, CCW-Net shapes a model's concept representations such that they are both necessary and sufficient for the outputs they are credited with and independent of other concepts' contributions. Counterfactual concept representations anchor a set of causal alignment criteria: necessity, sufficiency, and independence. Necessity drives irrelevant concept effects toward zero, sufficiency ensures relevant concept representations fully reconstruct a model's output, and independence enforces the total effect decomposes linearly across concepts. We evaluate CCW-Net across aircraft control, driving, and image classification. CCW-Net provides a step toward concept-based explanations that are not merely interpretable, but quantifiably causal, supporting safer evaluation, certification, and deployment of transparent and trustworthy neural models.

\end{abstract}

\section{Introduction}

Learned policies in autonomous systems are moving from research labs into high-stakes, real-world settings such as aviation \citep{pope2022hierarchical, ward2023us}, driving \citep{phan2022driving}, and robotics \citep{tang2025deep}. This is also the case for perception models in safety-relevant settings such as medical imaging. In task and safety-critical domains such as these, interpretability is becoming a necessary component \citep{rudin_stop_2019, atakishiyev2025safety}. Developers need it to troubleshoot \citep{kenny2026explainable}, testers need it to verify \citep{rountree2021testing, mahmud2024verification}, and users need it to understand \citep{sanneman2022situation}. Yet many neural models remain black boxes: they are not interpretable in a way that is useful for people in the task at hand. 

Concept-based interpretability offers promise by structuring decisions through human-interpretable concepts \citep{koh2020concept,echterhoff2024driving}. However, the link between concepts and outputs is typically correlational, not causal. A concept can activate alongside an output without being in the causal chain \citep{zhou2022feature}. An autonomous vehicle policy might activate a ``passenger pickup'' concept that has no causal bearing on the observed action. A bird classifier may activate ``has red plumage'' on every image of a cardinal while its prediction is actually driven by ``has crest''. The explanation appears correct but may mislead. To be useful for troubleshooting, certifying, and auditing, explanations must capture causal effects, i.e., the contribution of each concept to the output, isolated from concepts it co-occurs with \citep{pearl2012causal}. Correlation, no matter how sophisticated, does not address such causal questions \citep{pearl2009causality}. For safety and task-critical deployment we must transition from correlation to causation.

Beyond quantifying causal effects, some concept representations themselves limit what explanations can convey. Prior work typically represents concepts as scalars \citep{madumal2020explainable, koh2020concept, kenny2026explainable}, e.g., a ``left turn'' in driving or a ``has yellow belly'' attribute in bird classification. Scalars indicate the presence or strength of a concept but cannot capture their higher-dimensional, context-dependent expressions. We extend concept representations from scalars to vectors, enabling each concept to encode both \textit{whether} it exists and \textit{how} it is expressed.

We address these challenges with Causal Concept Wrapper Network (CCW-Net), the first use of mediation analysis as a training objective in deep networks. CCW-Net introduces a network architecture and adapts interventional multi-mediator mediation analysis from economics and epidemiology~\citep{vansteelandt2017interventional,loh2022disentangling,celli2022causal,lee2021guideline} to shape a model's concept representations so that each concept's contribution to the output is locally necessary, sufficient, and independent of contributions from other concepts. Counterfactuals required by these criteria are drawn from a mediator model (in our experiments, a per-concept invertible flow) trained alongside the overall model to track the conditional distribution of its concept embeddings. We further apply mediation analysis as an evaluation tool, quantifying, for each concept label combination, the share of a model's output causally attributable to each concept. This is finer-grained than the population-level alignment scores reported in prior concept-based work. We evaluate CCW-Net across aircraft control, driving, and fine-grained image classification, demonstrating that mediation-trained concepts produce explanations that are not merely interpretable, but quantifiably causal.

\paragraph{Main contributions.} Our contributions are threefold:

\begin{enumerate}[align=left,leftmargin=*,widest=9]
\item \textbf{Training objective for causal concepts in deep networks:} An interventional mediation analysis based objective that shapes each concept's causal effect on a model's output so that activated concepts are those causing the output thereby improving explanation trustworthiness.  
\item \textbf{Network architecture for concept-level mediation analysis:} A network architecture that drives absent concept representations toward zero supporting well-defined interventional mediation analyses and related training objectives (App.~\ref{sec:per-slot-identifiability}).
\item \textbf{Evaluation method for concept-based causality claims:} For any concept-bottleneck model, a method to quantify, for each concept label combination, how much a change in the model's output is causally attributable to each concept (App.~\ref{sec:estimator-details}).
\end{enumerate}

\section{Related work}
\label{sec:related-work}

\paragraph{Concept-based explanations.}
Concept-based interpretability is widely explored \citep{kim2018interpretability,AlvarezMelis2018towards,koh2020concept,bai2023concept,achtibat2023attribution,tan2024explain}. Concept Bottleneck Models \citep{koh2020concept} and adaptations to sequential decision-making \citep{kenny2023towards,kenny2026explainable,delfosse2024interpretable,das2023state2explanation} structure decisions through human-interpretable concepts but typically lack formal causal analysis. 

\paragraph{Causal mediation analysis.}
Causal mediation analysis models systems as directed acyclic graphs and decomposes an input's effect on an outcome into pathways through intermediate variables called mediators. Foundational tools have been developed in economics, medicine, and epidemiology \citep{mackinnon2008introduction,imai2010general,vanderweele2015explanation,pearl2012causal,celli2022causal,lee2021guideline}. Multi-mediator extensions \citep{vansteelandt2017interventional, loh2022disentangling} decompose total effects across multiple mediators without requiring causal ordering or independence assumptions among them, with indirect effects quantifying how much of an effect operates through a specific mediator. These tools are designed to estimate causal effects from observational data under strict assumptions (App. \ref{sec:causal-assumptions}), but to date have been applied to analyze causal relationships in data collected outside randomized controlled trials or trained models post-hoc, not as training objectives that shape deep networks' representations during learning.

\paragraph{Causal and disentanglement-based explanations for deep networks.}
A growing body of work applies causal frameworks to trained deep networks. \citet{vig2020investigating} applied mediation analysis post hoc to study bias in transformer language models, treating individual neurons and attention heads as mediators. \citet{geiger2021causal,geiger2025causal} formalize causal abstraction as a way to validate that a trained network implements a higher-level causal model. In concept-based interpretability specifically, recent work imposes or discovers causal structure over learned concepts \citep{de2026causally,dominici2025causal,qian2024learning}. Disentanglement methods like $\beta$-TCVAE \citep{chen2018isolating} and iVAE \citep{khemakhem2020variational} learn factorized representations evaluated by DCI \citep{eastwood2018framework} and FactorVAE score \citep{kim2018disentangling}. Across these methods, causal or representational analysis is applied post hoc to a model trained with conventional objectives, and disentanglement is a representational property that need not imply causal alignment \citep{scholkopf2021toward}. CCW-Net differs in that causal relationships between concept representations and model outputs are shaped during training rather than audited post-hoc. 

\paragraph{Feature attribution.}
Saliency-based \citep{simonyan2014deep,selvaraju2017grad,smilkov2017smoothgrad,sundararajan2017axiomatic} and perturbation-based \citep{ribeiro2016why,lundberg2017unified} attribution methods explain decisions by associating inputs or features to outputs. While useful for highlighting what features a model's outputs correspond to, these methods generally remain correlational. Ablating a feature changes a prediction without establishing causal responsibility \citep{adebayo2018sanity,zhou2022feature}. CCW-Net grounds causal attribution in interventional indirect effect estimates, yielding per-concept attributions with quantifiable causal estimates.

\section{Problem formulation and preliminaries}
\label{sec:problem}

We seek to train an interpretable policy that achieves expert-level task performance, provides concept-based explanations for its predicted outputs $\hat{Y}$, and ensures concept--output relationships reflect causal effects rather than spurious correlations. We assume access to a frozen pretrained backbone $f$ that achieves good task performance but lacks interpretability, and a dataset $\mathcal{D} = \{(X_i, Y_i, c_i)\}_{i=1}^N$ where $X_i \in \mathcal{X}$ is an input, $Y_i \in \mathcal{Y}$ is the expert output, and $c_i \in \{0,1\}^J$ is a vector of $J$ human-interpretable concept labels. We assume the concept ontology captures the key factors driving the model's outputs, the dataset provides sufficient ground truth, and standard ignorability conditions hold for the induced concept-to-output structural causal model (App.~\ref{sec:causal-assumptions}). We jointly optimize task fidelity, concept correctness, and causal alignment (Sec.~\ref{sec:method}), with concept--output causal relationships quantified in evaluation.

\paragraph{Structural causal models.} A structural causal model (SCM) \citep{pearl2009causality,peters2017elements} consists of equations $X_i = k_i(\mathrm{pa}_i, u_i)$ where each equation represents a mechanism determining exactly one variable, with $u_i$ the exogenous error term and $\mathrm{pa}_i$ the endogenous parent variables that directly cause $X_i$.

\paragraph{Interventional mediation analysis.} Mediation analysis \citep{pearl2012causal} decomposes the causal effect of a treatment on an outcome $Y$ into pathways that operate through variables called \emph{mediators} $M$. For mediators $M = (M_1, \dots, M_J)$, the total effect $\mathrm{TE}$ decomposes as $\mathrm{TE} = \mathrm{DE} + \mathrm{IE}$ into direct effect ($\mathrm{DE}$) and indirect effect ($\mathrm{IE}$) components. With multiple mediators, the indirect effect further decomposes per mediator $j$ as $\mathrm{IE} = \sum_j \mathrm{IE}_j$. Following \citet{loh2022disentangling}, we relax the assumption of causal ordering or independence among mediators by introducing a residual interaction term $\mathrm{IE}_\mu$, yielding $\mathrm{TE} = \mathrm{DE} + \sum_j \mathrm{IE}_j + \mathrm{IE}_\mu$. This relaxation is important for deep networks where operations like softmax and cross-attention induce dependencies among concept activations, which $\mathrm{IE}_\mu$ absorbs without violating identifiability. The \emph{interventional indirect effect} $\mathrm{IE}_j$ \citep{vansteelandt2017interventional} quantifies how much of the total effect operates through mediator $j$ when its representation $M_j$ is set to a counterfactual value while other mediator representations are held fixed. These quantities are identifiable from observational data under standard ignorability conditions \citep{vanderweele2015explanation,vansteelandt2017interventional,loh2022disentangling}, which hold by construction in the wrapper SCM (App.~\ref{sec:causal-assumptions}).

\section{Causal Concept Wrapper Network (CCW-Net)}
\label{sec:method}

CCW-Net wraps a frozen backbone policy with a concept encoder and new policy head which are trained to imitate the backbone's task performance as well as classify human-interpretable concepts similar to \citet{kenny2026explainable} but using a different architecture. In addition, training involves three causal alignment criteria to induce causal relationships between concepts and outputs: necessity, sufficiency, and independence. These criteria are then evaluated via interventional mediation analysis on the wrapper's outputs. Each causal alignment loss compares the head's output before and after one change of a training input's resulting concept $j$'s representation. In mediation analysis terms, the concept representations are mediators and the policy head is the outcome model. Every effect discussed is the model's own concept-to-output effect within the wrapper architecture's SCM, not a claim about inputs or observations to output effects. For our experiments, counterfactual concept representations required by the necessity alignment term are drawn from a per-concept invertible flow trained on the wrapper's encoder concepts. Any suitable modeling approach may be used. Reported evaluation metrics are produced from a separate estimator that does not use the mediator model. 

\subsection{Architecture}
\label{sec:method-arch}

Given a frozen pretrained backbone $f: \mathcal{X} \to \mathcal{Z}$ where $\mathcal{Z}$ is the frozen backbone's last latent layer, a concept encoder $g_\theta: \mathcal{Z} \to \mathbb{R}^{J \times D}$ maps $Z = f(X)$ to a concept representation $M = (M_1, \dots, M_J)$. Each $M_j \in \mathbb{R}^D$ represents the $j$-th human-interpretable concept. An attention head $h_\phi: \mathbb{R}^{J \times D} \to \mathcal{Y}$ maps the concept representation to a wrapper output $\hat{Y} = h_\phi(M)$. The wrapper induces the SCM $X \to Z \to M \to \hat{Y}$ with no direct $Z \to \hat{Y}$ edge so the input reaches the output only through the concept representations $M_j$ (DE$=$0), which gives mediator-outcome ignorability by construction (App.~\ref{sec:causal-assumptions}, Prop.~\ref{prop:ignorability}). The concept encoder $g_\theta$ is a multi-layer perceptron producing $J$ concept representation vectors of dimension $D$, each passed through a squash nonlinearity \citep{sabour2017dynamic} mapping each slot length onto $[0,1)$. The concept classification loss trains this length as presence leaving the direction free to encode the concept's context-based expression.

The concept classification loss drives inactive and active representations toward length zero and one, respectively. A zero concept vector representation carries a zero value vector in the attention head, so zeroing a concept removes its value contribution to $\hat{Y}$ regardless of its attention weight (Prop.~\ref{prop:per-slot}).

\begin{figure}
 \centering
 \includegraphics[width=1\linewidth]{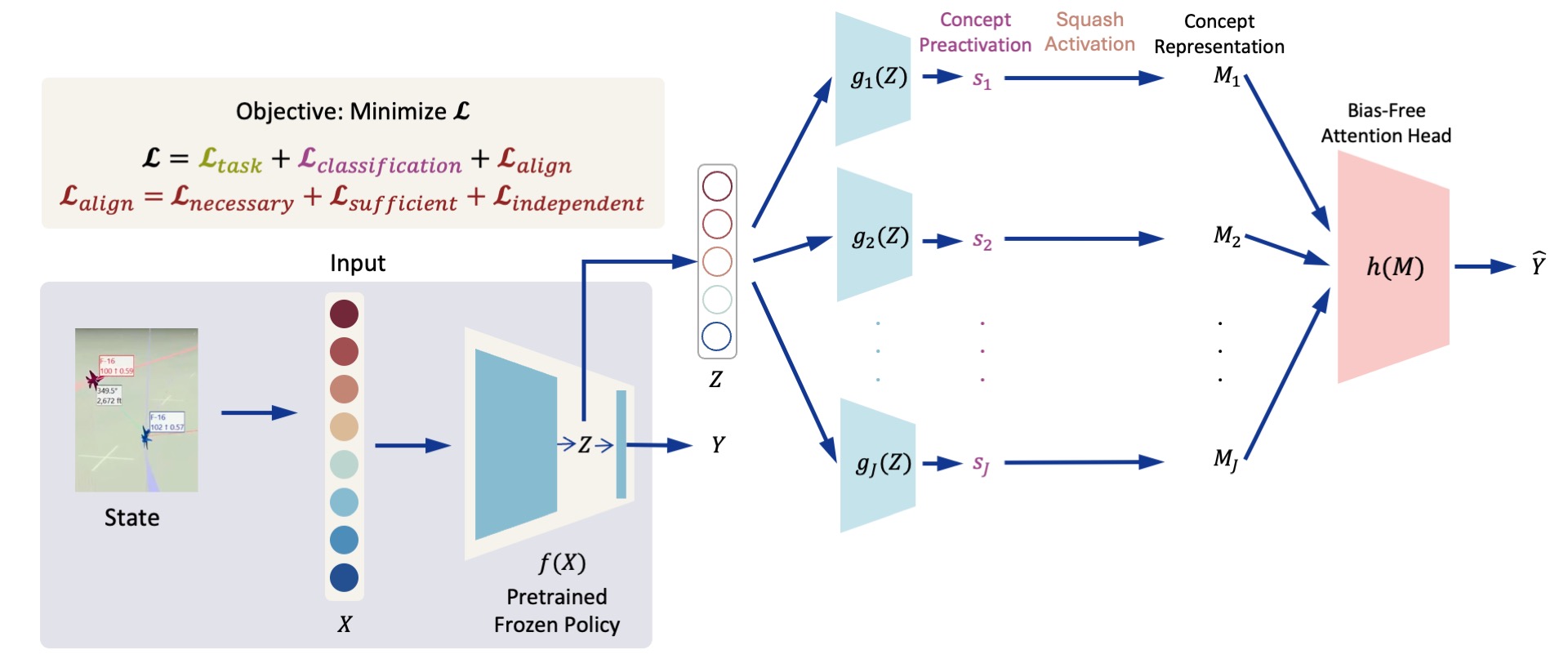}
\caption{\textbf{CCW-Net architecture.} A concept encoder $g_\theta$ maps the frozen backbone's latent $Z$ to one squashed vector $M_j$ per concept, whose length encodes the concept's presence and whose direction encodes its context-dependent expression, and a bias-free attention head $h_\phi$ maps $M$ to the wrapper output $\hat{Y}$. Causal alignment losses (weights and the auxiliary presence term omitted) are added to the task and concept losses, so that training shapes, and evaluation measures, how much each concept causally drives $\hat{Y}$.}
\end{figure}

\subsection{Mediator model: Counterfactual concepts via a per-concept flow}
\label{sec:method-flow}

Necessity asks whether the lack of a given concept's representation results in a lack of that concept manifesting in the model's outcome. In interventional mediation analysis, this is tested by replacing an inactive concept representation with an active one for the same input case. By definition, this is an unobservable counterfactual case. To overcome this, we reflect the approach of \citet{vansteelandt2017interventional} and draw counterfactual mediator expressions from a mediator model learned from observational data. In our experiments, we invoke a per-concept invertible flow $\mathcal{F}_\psi$ to model mediator distributions $p(M_j|c_j,X)$ and draw counterfactual examples where the concept label is flipped between present and not-present, 
\begin{equation}
\label{eq: counterfactual draw}
\tilde{M}_j \sim \mathcal{F}_\psi(\,\cdot\, \mid 1 - c_j, X).
\end{equation}
We fit the flow by maximum likelihood to the encoder's mediator representations after a warmup (Sec. \ref{sec:training objective}, App. \ref{sec:training-details}, Algorithm \ref{alg:ccw-net}). Our method is not specific to flow modeling. Any modeling approach may be chosen that best serves the domain. 

\subsection{Causal alignment criteria}
\label{sec:method-criteria}

Each causal alignment criterion (necessity, sufficiency, independence) is stated in terms of a change in the head output caused by an edit of a concept vector. Let $\mathbf{0}$ be the all-zero concept vector, $\mathbf{0}_{j \leftarrow M_j}$ the all-zero vector with slot $j$ set to its realized value $M_j$, and $M_{j \leftarrow \tilde{M}_j}$ the realized concepts $M$ with slot $j$ replaced by its counterfactual draw $\tilde{M}_j$ (Eq.~\ref{eq: counterfactual draw}). We define three changes:

\begin{equation}
\Delta^{0} = \hat{Y} - h_\phi(\mathbf{0}), \qquad
\Delta^{0}_j = h_\phi(\mathbf{0}_{j \leftarrow M_j}) - h_\phi(\mathbf{0}), \qquad
\Delta^{\mathrm{cf}}_j = h_\phi(M_{j \leftarrow \tilde{M}_j}) - \hat{Y}.
\end{equation}

$\Delta^{0}$ switches every concept slot on from $\mathbf{0}$ to its realized value, $\Delta^{0}_j$ switches slot $j$ alone on from $\mathbf{0}$, and $\Delta^{\text{cf}}_j$ swaps slot $j$ for its counterfactual draw, keeping all other slots at their realized values. The subscript indicates the edited slot, and the superscript indicates the edit: switching slots on from $\mathbf{0}$ (0) or swapping in a counterfactual draw from the mediator model (cf). In mediation analysis terms, the concept label flip in place of a treatment change gives $\Delta^{\mathrm{cf}}_j$ as the per-sample analog of the interventional indirect effect $\text{IE}_j$ of Sec.~\ref{sec:problem} and $\Delta^{0}_j$ the analogous contrast of concept slot $j$ against zero with the other slots at zero (App. \ref{sec:vd-connection}). Sec.~\ref{sec:experiments-eval} provides the estimator for the reported effects. 

Each causal alignment loss is based on these quantities grounded on dataset labels $c_j$ rather than on predicted concept presence. Below, we outline each causal alignment term with their formulations in Eq. \ref{eq:align-losses}.  

\paragraph{Necessity.} Inactive concepts should have no effect on the model's output: Replacing an inactive concept slot $j$ with a counterfactual should not change the output such that $\Delta^{\mathrm{cf}}_j=0$. 

\paragraph{Sufficiency.} Active concepts alone should produce the expert output: Starting from the all-zero output and adding each active concept's separate contribution should produce the expert output, $h_\phi(\mathbf{0}) + \textstyle\sum_{j:\, c_j = 1} \Delta^{0}_j = Y$. 

\paragraph{Independence.} Concept representations should contribute additively: Each concept should contribute on its own, and the per-concept contributions should sum to the total, $\Delta^{0}=\textstyle\sum_{j}\Delta^{0}_j$. 

\paragraph{Presence (auxiliary).} Since sufficiency constrains only the sum of concept representations, a subset of active concepts could carry all of the effect. To guard against that, we add an optional term that requires each active concept to carry at least an equal share $\beta=1/\textstyle\sum_{j}c_j$ of the total, $\|\Delta^{0}_j\|\geq\beta\|\Delta^{0}\|$ whenever $c_j=1$.

\begin{equation}
\label{eq:align-losses}
\begin{aligned}
\mathcal{L}_{\text{nec}} &= \mathbb{E}\Big[\textstyle\sum_{j:\, c_j = 0} \|\Delta^{\mathrm{cf}}_j\|^2\Big], &
\mathcal{L}_{\text{suf}} &= \mathbb{E}\Big[\big\|h_\phi(\mathbf{0}) + \textstyle\sum_{j:\, c_j = 1} \Delta^{0}_j - Y\big\|^2\Big], \\
\mathcal{L}_{\text{ind}} &= \mathbb{E}\Big[\big\|\Delta^{0} - \textstyle\sum_{j} \Delta^{0}_j\big\|^2\Big], &
\mathcal{L}_{\text{pres}} &= \mathbb{E}\Big[\textstyle\sum_{j:\, c_j = 1} \max\big(0,\, \beta\|\Delta^{0}\| - \|\Delta^{0}_j\|\big)^2\Big].
\end{aligned}
\end{equation}

\subsection{Training objective}
\label{sec:training objective}

We combine the four alignment losses into a single alignment term
\begin{equation}
\mathcal{L}_{\text{align}} = \lambda_{\text{nec}} \mathcal{L}_{\text{nec}} + \lambda_{\text{suf}} \mathcal{L}_{\text{suf}} + \lambda_{\text{ind}} \mathcal{L}_{\text{ind}} + \lambda_{\text{pres}} \mathcal{L}_{\text{pres}},
\end{equation}
and train the wrapper by minimizing
\begin{equation}
\mathcal{L}_{\text{wrap}}(\theta, \phi) = \mathcal{L}_{\text{task}} + \lambda_{\text{cls}} \mathcal{L}_{\text{cls}} + \mathcal{L}_{\text{align}}.
\end{equation}
$\mathcal{L}_{\text{task}}$ is the supervised mean-squared error on $\hat{Y}$ (to the expert action for control, to the one-hot species label for CUB-200) and $\mathcal{L}_{\text{cls}}$ trains each concept slot's vector length to predict its concept label. Training proceeds in three phases (Algorithm~\ref{alg:ccw-net}): warmup on $\mathcal{L}_{\text{task}}$ and $\mathcal{L}_{\text{cls}}$, an intermission that fits the mediator model on the frozen encoder's concepts, and alignment on the full objective, with optional mediator model refits at fixed intervals. Loss weights, schedules, and architecture details are in Apps.~\ref{sec:architecture-details}--\ref{sec:training-details}.

\section{Empirical evaluation}
\label{sec:experiments}

We evaluate CCW-Net on three domains spanning control and perception: aircraft control, the CarRacing-v1 driving environment \citep{brockman2016openai}, and image classification on CUB-200 \citep{wah2011caltech}.

\subsection{Tasks}

\paragraph{Aircraft control.}
We evaluate on extended trail formation \citep{aetcman11-251-2024,aetcman11-248-2025}, in which a chase aircraft maintains position behind a lead aircraft executing dynamic maneuvers. The simulator is a high-fidelity aircraft model with six degrees of freedom \citep{heidlauf2018verification} where the lead aircraft follows a scripted trajectory and the chase aircraft is governed by a reinforcement learning policy trained by proximal policy optimization \citep{schulman2017proximal} using the aircraft environment and codebase of \citet{So-RSS-23}. The action space is two-dimensional (pitch and roll commands). Concept labels are $J = 4$ binary attributes (\emph{Lead pursuit}, \emph{Lag pursuit}, \emph{Climb}, \emph{Dive}) drawn from operational pilot vocabulary \citep{aetcman11-251-2024,aetcman11-248-2025}. Pursuit (\emph{Lead pursuit}, \emph{Lag pursuit}) and vertical (\emph{Climb}, \emph{Dive}) maneuver concept pairs are mutually exclusive, so exactly two concepts are active per sample. Full task details are in Appendix~\ref{sec:aircraft-task}.

\paragraph{CarRacing.}
We evaluate on the CarRacing-v1 environment \citep{brockman2016openai}, in which a pixel-input agent drives a car around a track. The frozen backbone is a CNN policy trained end-to-end on the task. Its latents, actions, and concept labels come from a CW-Net test domain \citep{kenny2026explainable}. The action space is continuous steering, throttle, and brake. Concept labels are $J=3$ binary attributes (\emph{Left}, \emph{Right}, \emph{Straight}). The concepts are mutually exclusive and exhaustive, so exactly one concept is active per sample and the presence threshold is $\beta=1$. Full task details are in Appendix~\ref{sec:carracing-task}.

\paragraph{CUB-200 classification.}
We evaluate on the CUB-200-2011 dataset of 200 bird species \citep{wah2011caltech}. The frozen backbone is an Inception-v3 image classifier fine-tuned on CUB. Concept labels are $J = 112$ binary visual attributes (e.g., ``has yellow belly'', ``has pointed beak'') and the wrapper output is a 200-way classification. Full task details are in Appendix~\ref{sec:cub-task}.

\subsection{Baselines}

\paragraph{Same-architecture, no alignment.}
We compare CCW-Net against the same encoder and head trained without the alignment losses (only $\mathcal{L}_{\text{task}}$ and $\mathcal{L}_{\text{cls}}$ active). This isolates the contribution of the causal alignment objective from the contribution of the encoder architecture.

\paragraph{Concept Wrapper Network variants.}
On the aircraft and CarRacing tasks we compare against the Concept Wrapper Network (CW-Net) of \citet{kenny2026explainable}, using the authors' architecture. CW-Net wraps a frozen backbone with a concept layer in the same spirit as CCW-Net, but implements a different architecture, represents each concept as a scalar, and trains without causal alignment. Appendix~\ref{sec:appendix-ablations} adds CW-Net variants trained with alignment and, on CarRacing, with vector concepts.

\paragraph{Concept bottleneck model.}
For CUB-200 we compare against the concept bottleneck model of \citet{koh2020concept}, which trains a model to predict concepts and to derive the task output through them.

\paragraph{Disentanglement-based concept learning.}
To support the secondary claim that disentanglement does not imply causal alignment, we compare against $\beta$-TCVAE \citep{chen2018isolating} and iVAE \citep{khemakhem2020variational} on the aircraft and CarRacing tasks, reporting DCI \citep{eastwood2018framework} alongside the concept effect fractions (Appendix~\ref{sec:appendix-ablations}).

\subsection{Ablations}
\label{sec:experiments-ablations}

We run five ablation types on the control domains: (i) \textbf{loss combinations} where subsets of the four alignment losses are disabled to test each one's contribution, (ii) \textbf{concept representation dimension} with $D = 16$ default, $D = 1$ scalar, and $D = 8$, (iii) \textbf{data quantity} for aircraft ranging from 5K to 5M samples, (iv) \textbf{label quality} as fraction of data labeled with concepts and injected label noise, and (v) \textbf{concept ontology} where we remove concepts, merge concepts, and add a placebo concept. The full ablation matrix is in Appendix~\ref{sec:appendix-ablations}. The CUB-200 models train without the alignment losses and compare architectures (Sec.~\ref{sec:experiments-cub}).

\subsection{Evaluation methodology}
\label{sec:experiments-eval}

We report task performance (mean-squared error for action regression), concept accuracy against data labels, DCI \citep{eastwood2018framework}, and the causal effect each concept has on the model's output. All values are means over ten seeds with two-sided 95\% Student-$t$ intervals. The primary metrics in this study are causal effects. 

\paragraph{Causal attribution.}
To test whether a model's concept representations carry only their own effects, we estimate interventional indirect effects \citep{vansteelandt2017interventional, loh2022disentangling} from held-out observational data, using no mediator model, no zero reference $\mathbf{0}$, and none of the training losses. The approach applies to any model whose output is a function of its concept representations. We use it on CCW-Net, CW-Net, and the disentanglement baselines.

Each comparison takes two sets of test inputs, $\mathcal{A}$ and $\mathcal{B}$, whose concept labels differ where an active concept in $\mathcal{A}$ is replaced in $\mathcal{B}$ by a mutually exclusive alternative with all other labels unchanged. For a test input $i$, we draw matches from $\mathcal{A}$ and $\mathcal{B}$ by Gaussian-kernel matching on z-scored features $z$ (raw inputs or for image inputs, backbone features $f(X)$). A test input $i' \neq i$ is sampled with probability proportional to $\exp(-d_{ii'}^2/2\tau^2)$ among those within a radius of $i$, where $d_{ii'}$ is the root-mean-square distance between $z_i$ and $z_{i'}$ and $\tau$ is the kernel bandwidth. If none are within the radius, we take the nearest neighbor. The matches' concepts $g_\theta(f(X_{i'}))$ fill the concept slots. The two slots whose labels differ form one group, so every hybrid has a label combination that occurs in the data, and every other slot is a group of its own. We then build hybrid concept vectors taking each group of concept mediator representation slots from the $\mathcal{A}$ or the $\mathcal{B}$ match, and pass them through the trained head $h_\phi$. Group $G$'s effect is the change in output when it switches from its $\mathcal{A}$ match to its $\mathcal{B}$ match, 
\begin{equation}
\label{eq:ifm}
\mathrm{IE}_G=\mathbb{E}_{i,\pi}\Big[h_\phi\big(M^{\mathcal{B}}_{\pi<G},\,M^{\mathcal{B}}_{G},\,M^{\mathcal{A}}_{\pi>G}\big)-h_\phi\big(M^{\mathcal{B}}_{\pi<G},\,M^{\mathcal{A}}_{G},\,M^{\mathcal{A}}_{\pi>G}\big)\Big],
\end{equation}
where $M^{\mathcal{A}}_G$ ($M^{\mathcal{B}}_G$) are group $G$'s slots from an $\mathcal{A}$ ($\mathcal{B}$) match, drawn independently for each group, and likewise for the groups $\pi<G$ ($\pi>G$) before (after) $G$ in an ordering $\pi$. Averaging over all orderings makes $\mathrm{IE}_G$ group $G$'s Shapley value for the switch from $\mathcal{A}$ to $\mathcal{B}$, and $i$ runs over the inputs of both sets. Estimator details are in App.~\ref{sec:estimator-details}.

For each label combination (rows in Tables~\ref{tab:aircraft-per-state} and~\ref{tab:carracing-per-state}), slot $j$'s effect $\|\mathrm{IE}_j\|$ is the norm of its own group's effect in the comparisons that keep $j$'s label fixed, i.e., how far slot $j$ moves the output when another concept changes. The fractions of the effect that slot $j$, the active concepts, and the inactive concepts carry are
\begin{equation}
\label{eq:rho}
\rho_j = \frac{\|\mathrm{IE}_j\|}{\sum_k \|\mathrm{IE}_k\|}, \qquad
\rho_{\text{active}} = \sum_{j:\, c_j = 1} \rho_j, \qquad
\rho_{\text{inactive}} = \sum_{j:\, c_j = 0} \rho_j = 1 - \rho_{\text{active}}.
\end{equation}
Inactive concepts should carry no effect, so a faithful model has $\rho_{\text{inactive}}$ near zero.

On CUB-200, we report a separate per-input evaluation. Its 112 attribute labels are assigned per species and have no mutually exclusive alternatives within the dataset, so the comparisons above, which hold all other labels fixed, cannot be formed with enough test inputs. Instead, each concept in turn is set to absent (the zero slot for CCW-Net, concept probability $0$ for CBM), $\rho_j$ is its share of the summed output changes at each test input, and we average over the test split. This uses the zero reference, but no CUB-200 model is trained with the alignment losses, so it reuses no optimized terms.

\subsection{Results}
\label{sec:experiments-results}

\subsubsection{Aircraft control}
\label{sec:experiments-aircraft}

Table~\ref{tab:aircraft-per-state} reports, for each concept label combination, the fraction of the effect each concept representation mediator carries. CCW-Net (concept representation dimension $D = 16$) routes $94.5 \pm 2.2\%$ of the effect through the active concepts, versus $88.1 \pm 4.3\%$ trained without the causal alignment losses maintaining comparable task error and concept accuracy. CW-Net predicts its concepts nearly as accurately ($91.2\%$ and $98.2\%$ on the two mutually exclusive concept sets), yet its active concepts carry only $50.0 \pm 6.2\%$ of the effect. This is the fraction expected if the effect were spread indiscriminately across the four concepts. This suggests accurate concept prediction alone does not make the model's outcomes depend on them. We find DCI disentanglement is near its ceiling of $1$ for every model in Table~\ref{tab:aircraft-per-state} suggesting that disentanglement metrics may be insufficient to imply causal relationships between representation and model outcomes. The disentanglement baseline iVAE reaches DCI disentanglement, completeness and informativeness of $1.000$, $0.891$ and $1.000$, yet its active concepts carry only $47.6 \pm 2.4\%$ of the effect (App.~\ref{sec:aircraft-full-ablations}).

\begin{table}[t]
\centering
\caption{Aircraft control: fraction of the effect each concept carries (Eq.~\ref{eq:rho}), per concept label combination, estimated as in Sec.~\ref{sec:experiments-eval}. Active concepts are in bold. All Active and All Inactive are $\rho_{\text{active}}$ and $\rho_{\text{inactive}}$. Each block's header gives the model's test action MSE (tMSE), concept accuracy on the Lead/Lag and Climb/Dive concept sets (\%), and DCI disentanglement / completeness / informativeness. Means over ten seeds with two-sided 95\% Student-$t$ intervals. CCW-Net shows superior causal effect routing through active concepts compared to prior art.}
\label{tab:aircraft-per-state}
\small
\setlength{\tabcolsep}{4pt}
\begin{tabular}{lrrrrrr}
\toprule
 & \multicolumn{6}{c}{Concept Effect (\%)} \\
\cmidrule(lr){2-7}
Label combination & Lead & Lag & Climb & Dive & All Active $\uparrow$ & All Inactive $\downarrow$\\
\midrule
\multicolumn{7}{l}{\textit{CCW-Net (Ours)}} \\
\multicolumn{7}{l}{\footnotesize tMSE 0.125\,{\scriptsize$\pm$0.001}; concept acc.\ 95.7\,{\scriptsize$\pm$0.1} / 99.3\,{\scriptsize$\pm$0.0}; DCI 0.999\,{\scriptsize$\pm$0.001} / 0.719\,{\scriptsize$\pm$0.041} / 0.975\,{\scriptsize$\pm$0.001}} \\
Lead+Climb & \textbf{62.8}\,{\scriptsize$\pm$16.1} & 4.6\,{\scriptsize$\pm$1.9} & \textbf{32.0}\,{\scriptsize$\pm$16.1} & 0.6\,{\scriptsize$\pm$0.4} & 94.8\,{\scriptsize$\pm$2.1} & 5.2\,{\scriptsize$\pm$2.1} \\
Lead+Dive & \textbf{69.2}\,{\scriptsize$\pm$15.7} & 5.4\,{\scriptsize$\pm$2.9} & 1.0\,{\scriptsize$\pm$0.5} & \textbf{24.4}\,{\scriptsize$\pm$16.0} & 93.6\,{\scriptsize$\pm$3.3} & 6.4\,{\scriptsize$\pm$3.3} \\
Lag+Climb & 4.3\,{\scriptsize$\pm$2.3} & \textbf{60.5}\,{\scriptsize$\pm$13.2} & \textbf{34.5}\,{\scriptsize$\pm$13.5} & 0.7\,{\scriptsize$\pm$0.4} & 95.0\,{\scriptsize$\pm$2.5} & 5.0\,{\scriptsize$\pm$2.5} \\
Lag+Dive & 4.9\,{\scriptsize$\pm$2.4} & \textbf{67.8}\,{\scriptsize$\pm$13.0} & 1.1\,{\scriptsize$\pm$0.6} & \textbf{26.2}\,{\scriptsize$\pm$13.3} & 94.0\,{\scriptsize$\pm$2.6} & 6.0\,{\scriptsize$\pm$2.6} \\
\midrule
\multicolumn{7}{l}{\textit{CCW-Net, no causal alignment losses}} \\
\multicolumn{7}{l}{\footnotesize tMSE 0.124\,{\scriptsize$\pm$0.001}; concept acc.\ 95.7\,{\scriptsize$\pm$0.1} / 99.3\,{\scriptsize$\pm$0.0}; DCI 0.999\,{\scriptsize$\pm$0.000} / 0.746\,{\scriptsize$\pm$0.037} / 0.975\,{\scriptsize$\pm$0.001}} \\
Lead+Climb & \textbf{63.7}\,{\scriptsize$\pm$21.0} & 9.5\,{\scriptsize$\pm$6.0} & \textbf{25.0}\,{\scriptsize$\pm$16.3} & 1.8\,{\scriptsize$\pm$2.0} & 88.7\,{\scriptsize$\pm$7.2} & 11.3\,{\scriptsize$\pm$7.2} \\
Lead+Dive & \textbf{64.8}\,{\scriptsize$\pm$20.0} & 9.6\,{\scriptsize$\pm$5.6} & 4.1\,{\scriptsize$\pm$3.9} & \textbf{21.5}\,{\scriptsize$\pm$14.0} & 86.3\,{\scriptsize$\pm$6.8} & 13.7\,{\scriptsize$\pm$6.8} \\
Lag+Climb & 8.0\,{\scriptsize$\pm$2.8} & \textbf{62.6}\,{\scriptsize$\pm$16.9} & \textbf{27.3}\,{\scriptsize$\pm$15.6} & 2.0\,{\scriptsize$\pm$2.0} & 89.9\,{\scriptsize$\pm$3.7} & 10.1\,{\scriptsize$\pm$3.7} \\
Lag+Dive & 8.5\,{\scriptsize$\pm$4.3} & \textbf{63.4}\,{\scriptsize$\pm$17.9} & 5.2\,{\scriptsize$\pm$6.3} & \textbf{22.9}\,{\scriptsize$\pm$14.6} & 86.3\,{\scriptsize$\pm$6.3} & 13.7\,{\scriptsize$\pm$6.3} \\
\midrule
\multicolumn{7}{l}{\textit{CW-Net \citep{kenny2026explainable}}} \\
\multicolumn{7}{l}{\footnotesize tMSE 0.122\,{\scriptsize$\pm$0.002}; concept acc.\ 91.2\,{\scriptsize$\pm$0.2} / 98.2\,{\scriptsize$\pm$0.1}; DCI 0.998\,{\scriptsize$\pm$0.001} / 0.719\,{\scriptsize$\pm$0.064} / 0.947\,{\scriptsize$\pm$0.001}} \\
Lead+Climb & \textbf{30.5}\,{\scriptsize$\pm$11.8} & 41.0\,{\scriptsize$\pm$11.0} & \textbf{10.6}\,{\scriptsize$\pm$4.5} & 17.8\,{\scriptsize$\pm$10.6} & 41.2\,{\scriptsize$\pm$12.7} & 58.8\,{\scriptsize$\pm$12.7} \\
Lead+Dive & \textbf{25.2}\,{\scriptsize$\pm$11.2} & 31.4\,{\scriptsize$\pm$8.6} & 17.3\,{\scriptsize$\pm$6.6} & \textbf{26.0}\,{\scriptsize$\pm$9.8} & 51.3\,{\scriptsize$\pm$6.3} & 48.7\,{\scriptsize$\pm$6.3} \\
Lag+Climb & 31.5\,{\scriptsize$\pm$13.2} & \textbf{38.8}\,{\scriptsize$\pm$13.2} & \textbf{11.5}\,{\scriptsize$\pm$4.7} & 18.2\,{\scriptsize$\pm$8.4} & 50.3\,{\scriptsize$\pm$13.0} & 49.7\,{\scriptsize$\pm$13.0} \\
Lag+Dive & 26.1\,{\scriptsize$\pm$13.1} & \textbf{28.2}\,{\scriptsize$\pm$9.2} & 18.3\,{\scriptsize$\pm$6.5} & \textbf{27.4}\,{\scriptsize$\pm$8.5} & 55.6\,{\scriptsize$\pm$14.1} & 44.4\,{\scriptsize$\pm$14.1} \\
\bottomrule
\end{tabular}
\end{table}

\subsubsection{CarRacing}
\label{sec:experiments-carracing}

Table~\ref{tab:carracing-per-state} reports the same for CarRacing, where one of the three concepts is active in each label combination. CCW-Net (concept representation dimension $D=1$) routes $89.8 \pm 2.5\%$ of the effect through the active concept, versus $87.5 \pm 4.9\%$ trained without the causal alignment losses at the same task error and comparable concept accuracy, so on CarRacing the architecture accounts for most of the causal routing. While CW-Net reaches lower task error than CCW-Net, both policies were able to complete the task upon visual inspection. While both approaches predict their concepts nearly as accurately ($92.6 \pm 0.4\%$ and $92.4 \pm 0.4\%$ vs $90.0 \pm 1.3\%$) CW-Net's active concept carries only $33.1 \pm 1.6\%$ of the effect, the fraction expected if the effect were spread indiscriminately across the three concepts.

\begin{table}[t]
\centering
\caption{CarRacing: fraction of the effect each concept carries (Eq.~\ref{eq:rho}), per concept label combination, estimated as in Sec.~\ref{sec:experiments-eval}. Active concepts are in bold. All Active and All Inactive are $\rho_{\text{active}}$ and $\rho_{\text{inactive}}$. Each block's header gives the model's test action MSE (tMSE), concept accuracy on the Straight/Left/Right concept set (\%), and DCI disentanglement / completeness / informativeness. Means over ten seeds with two-sided 95\% Student-$t$ intervals. CCW-Net shows superior causal effect routing through active concepts compared to prior art.}
\label{tab:carracing-per-state}
\small
\setlength{\tabcolsep}{4pt}
\begin{tabular}{lrrrrr}
\toprule
 & \multicolumn{5}{c}{Concept Effect (\%)} \\
\cmidrule(lr){2-6}
Label combination & Straight & Left & Right & All Active $\uparrow$ & All Inactive $\downarrow$\\
\midrule
\multicolumn{6}{l}{\textit{CCW-Net (Ours)}} \\
\multicolumn{6}{l}{\footnotesize tMSE 0.0419\,{\scriptsize$\pm$0.0005}; concept acc.\ 92.6\,{\scriptsize$\pm$0.4}; DCI 0.863\,{\scriptsize$\pm$0.023} / 0.863\,{\scriptsize$\pm$0.023} / 0.943\,{\scriptsize$\pm$0.005}} \\
Straight & \textbf{98.1}\,{\scriptsize$\pm$1.1} & 1.3\,{\scriptsize$\pm$0.9} & 0.5\,{\scriptsize$\pm$0.4} & 98.1\,{\scriptsize$\pm$1.1} & 1.9\,{\scriptsize$\pm$1.1} \\
Left & 25.2\,{\scriptsize$\pm$7.0} & \textbf{74.5}\,{\scriptsize$\pm$7.0} & 0.3\,{\scriptsize$\pm$0.2} & 74.5\,{\scriptsize$\pm$7.0} & 25.5\,{\scriptsize$\pm$7.0} \\
Right & 23.5\,{\scriptsize$\pm$5.8} & 0.7\,{\scriptsize$\pm$0.4} & \textbf{75.8}\,{\scriptsize$\pm$5.8} & 75.8\,{\scriptsize$\pm$5.8} & 24.2\,{\scriptsize$\pm$5.8} \\
\midrule
\multicolumn{6}{l}{\textit{CCW-Net, no causal alignment losses}} \\
\multicolumn{6}{l}{\footnotesize tMSE 0.0419\,{\scriptsize$\pm$0.0006}; concept acc.\ 92.4\,{\scriptsize$\pm$0.4}; DCI 0.850\,{\scriptsize$\pm$0.021} / 0.851\,{\scriptsize$\pm$0.021} / 0.944\,{\scriptsize$\pm$0.005}} \\
Straight & \textbf{95.4}\,{\scriptsize$\pm$3.3} & 3.5\,{\scriptsize$\pm$2.5} & 1.2\,{\scriptsize$\pm$0.9} & 95.4\,{\scriptsize$\pm$3.3} & 4.6\,{\scriptsize$\pm$3.3} \\
Left & 26.2\,{\scriptsize$\pm$9.2} & \textbf{73.2}\,{\scriptsize$\pm$9.4} & 0.6\,{\scriptsize$\pm$0.3} & 73.2\,{\scriptsize$\pm$9.4} & 26.8\,{\scriptsize$\pm$9.4} \\
Right & 24.5\,{\scriptsize$\pm$7.5} & 1.7\,{\scriptsize$\pm$1.0} & \textbf{73.9}\,{\scriptsize$\pm$7.9} & 73.9\,{\scriptsize$\pm$7.9} & 26.1\,{\scriptsize$\pm$7.9} \\
\midrule
\multicolumn{6}{l}{\textit{CW-Net \citep{kenny2026explainable}}} \\
\multicolumn{6}{l}{\footnotesize tMSE 0.0015\,{\scriptsize$\pm$0.0007}; concept acc.\ 90.0\,{\scriptsize$\pm$1.3}; DCI 0.078\,{\scriptsize$\pm$0.063} / 0.626\,{\scriptsize$\pm$0.082} / 0.928\,{\scriptsize$\pm$0.008}} \\
Straight & \textbf{33.2}\,{\scriptsize$\pm$1.7} & 37.4\,{\scriptsize$\pm$7.4} & 29.4\,{\scriptsize$\pm$6.8} & 33.2\,{\scriptsize$\pm$1.7} & 66.8\,{\scriptsize$\pm$1.7} \\
Left & 33.7\,{\scriptsize$\pm$11.0} & \textbf{32.9}\,{\scriptsize$\pm$1.4} & 33.4\,{\scriptsize$\pm$10.4} & 32.9\,{\scriptsize$\pm$1.4} & 67.1\,{\scriptsize$\pm$1.4} \\
Right & 29.7\,{\scriptsize$\pm$7.0} & 37.3\,{\scriptsize$\pm$6.5} & \textbf{33.1}\,{\scriptsize$\pm$2.1} & 33.1\,{\scriptsize$\pm$2.1} & 66.9\,{\scriptsize$\pm$2.1} \\
\bottomrule
\end{tabular}
\end{table}

\subsubsection{CUB-200}
\label{sec:experiments-cub}

Table~\ref{tab:cub-per-bucket} reports CUB-200. Neither CCW-Net nor CBM is trained with the causal alignment losses, so this is a comparison that isolates the architecture. CCW-Net (concept representation dimension $D=1$) routes $91.3 \pm 0.3\%$ of the effect through active concepts, versus $41.6 \pm 0.6\%$ for CBM with both predicting their concepts accurately ($99.6 \pm 0.0\%$ and $97.7 \pm 0.1\%$). More than half of CBM's effect ($58.4 \pm 0.6\%$) is attributed to concepts labeled absent. Similar to the above studies, accurate concept prediction alone does not make the model's outcomes depend on the active concepts. In CCW-Net, a zeroed concept vector carries a zero key and zero value in the attention head, so setting a concept to absent removes its contribution to the outcome $\hat{Y}$ (Prop.~\ref{prop:per-slot}).

\begin{table}[t]
\centering
\caption{CUB-200: fraction of the effect the active concepts carry, per input (Sec.~\ref{sec:experiments-eval}), averaged over the test split. CBM is the concept bottleneck model of \citet{koh2020concept}. No CUB-200 model is trained with the causal alignment losses. Concept accuracy is over the 112 attributes. Means over ten seeds with two-sided 95\% Student-$t$ intervals.}
\label{tab:cub-per-bucket}
\small
\setlength{\tabcolsep}{2.5pt}
\begin{tabular}{lrrrrrr}
\toprule
 & Concept & \multicolumn{2}{c}{Concept Effect (\%)} & \multicolumn{3}{c}{DCI} \\
\cmidrule(lr){3-4}\cmidrule(lr){5-7}
Model & acc.\ (\%) & All Active $\uparrow$ & All Inactive $\downarrow$ & Disentanglement & Completeness & Informativeness \\
\midrule
CCW-Net (Ours) & 99.6\,{\scriptsize$\pm$0.0} & 91.3\,{\scriptsize$\pm$0.3} & 8.7\,{\scriptsize$\pm$0.3} & 0.895\,{\scriptsize$\pm$0.007} & 0.936\,{\scriptsize$\pm$0.007} & 0.993\,{\scriptsize$\pm$0.001} \\
CBM & 97.7\,{\scriptsize$\pm$0.1} & 41.6\,{\scriptsize$\pm$0.6} & 58.4\,{\scriptsize$\pm$0.6} & 0.732\,{\scriptsize$\pm$0.011} & 0.789\,{\scriptsize$\pm$0.011} & 0.985\,{\scriptsize$\pm$0.001} \\
\bottomrule
\end{tabular}
\end{table}

\subsubsection{Robustness and ablation}
\label{sec:experiments-robustness}

We performed ablation studies on the control domains: aircraft control and CarRacing. Full results are in Tables~\ref{tab:aircraft-full-ablations} and~\ref{tab:carracing-full-ablations} (App.~\ref{sec:appendix-ablations}).

\paragraph{Label efficiency.}
Across both control domains, CCW-Net retains most of its causal faithfulness when concept supervision is increasingly sparse. For example, when only 10\% of data is labeled with concepts, $\rho_{\text{active}}$ decreases by only $4.0$ points on aircraft control ($97.3\%$ to $93.3\%$) and interestingly increases by $2.6$ on CarRacing ($75.3\%$ to $77.9\%$). Both are within noise across seeds. We believe this is consistent with CCW-Net's alignment losses and architecture providing strong enough inductive bias that dense labeling becomes less critical. These results are encouraging as they suggest that less labeling burden is required to retain comparable causal alignment results. 

\paragraph{Label noise.}
We find that label noise has the opposite effect. 10\% noise drops $\rho_{\text{active}}$ from $97.3\%$ to $71.6\%$ on aircraft control and from $75.3\%$ to $61.8\%$ on CarRacing while concept accuracy remains nearly unchanged ($96.2\%$ vs.\ $96.4\%$ and $92.3\%$ vs.\ $93.4\%$). This is an example where concept accuracy and architecture alone do not reflect causal faithfulness motivating the need for evaluation approaches such as ours to verify such claims. 

\paragraph{Ontology misspecification.}
CCW-Net degrades gracefully when the ontology isn't fully specified. Removing one concept from the aircraft ontology lowers $\rho_{\text{active}}$ to 79.9--94.6\%, and merging two concepts leaves it at 95.6--97.4\%. On CarRacing, removing one concept gives 65.8--82.1\% and merging two gives 56.9--95.7\%, against $75.3\%$ for the full ontology.

\paragraph{Loss ablation.}
On aircraft control, removing the sufficiency or independence loss terms lowers $\rho_{\text{active}}$ from $97.3\%$ to $94.4\%$ and $94.8\%$, while removing the presence term does not ($98.0\%$). On CarRacing, where one concept is active per input, necessity alone reaches $77.2\%$, comparable to $75.3\%$ with all four losses, and presence alone reaches $71.9\%$ against $63.5\%$ without alignment. This suggests different domains benefit from the loss terms differently. 

\paragraph{Vector versus scalar concepts.}
We find concept representation dimension $D>1$ allows the alignment losses flexibility to improve causal alignment. At $D=16$, alignment increases $\rho_{\text{active}}$ by 8.1 points on aircraft control and 11.8 points on CarRacing, while at $D = 1$ the change is $-0.9$ and $2.4$ points.

\section{Conclusion}

Concept-based explanations are useful to the extent that they causally drive the model's outputs. We presented CCW-Net, the first formulation of multi-mediator interventional mediation analysis as a training objective for deep networks. We also introduced an evaluation method that quantifies the fraction of a model's output effect concept explanations carry. CCW-Net wraps a frozen backbone model with a concept encoder and attention head, trained to satisfy three causal alignment criteria (necessity, sufficiency, and independence) with counterfactual concept representations drawn from a mediator model. CCW-Net's causal claims concern the effect of concept activations and representations on model outcomes where effects in the data-generation process are left to future work. 

Across aircraft control, CarRacing-v1, and CUB-200 image classification, CCW-Net routes more of its output's effect through active concepts than concept bottleneck based approaches. The contribution combines two approaches: concept representations feeding an attention head in which absent concepts contribute zero value, isolating each concept's contribution to the output, and causal alignment losses that strengthen this effect and improve consistency across seeds. On our aircraft control domain, CCW-Net routes $94.5 \pm 2.2\%$ of the effect through active concepts against $50.0 \pm 6.2\%$ for prior art. On CUB-200, the architecture alone routes 91.3\% against 41.6\% for CBM at comparable concept accuracy, suggesting that concept accuracy is not an indicator of causal faithfulness. iVAE achieves excellent DCI disentanglement metrics on aircraft control, yet its active concepts carry only 47.6\% of the effect, suggesting disentanglement and causal alignment are different properties.

CCW-Net moves causal faithfulness from a property that is claimed by inspection to one that can be designed for, trained against, and measured by. Concept-based explanations become not merely interpretable but quantifiably causal thereby building a road toward verifiably trustworthy models in safety- and task-critical environments. 

\begin{ack}
Research was sponsored by the Department of the Air Force Artificial Intelligence Accelerator and was accomplished under Cooperative Agreement Number FA8750-19-2-1000. The views and conclusions contained in this document are those of the authors and should not be interpreted as representing the official policies, either expressed or implied, of the Department of the Air Force or the U.S. Government. The U.S. Government is authorized to reproduce and distribute reprints for Government purposes notwithstanding any copyright notation herein. The authors acknowledge the MIT SuperCloud and Lincoln Laboratory Supercomputing Center for providing HPC resources that have contributed to the research results reported within this paper. The authors declare no competing interests.
\end{ack}

\small
\bibliographystyle{plainnat}
\bibliography{neurips2026}

\begin{thebibliography}{80}
\providecommand{\natexlab}[1]{#1}
\providecommand{\url}[1]{\texttt{#1}}
\expandafter\ifx\csname urlstyle\endcsname\relax
  \providecommand{\doi}[1]{doi: #1}\else
  \providecommand{\doi}{doi: \begingroup \urlstyle{rm}\Url}\fi

\bibitem[Achtibat et~al.(2023)Achtibat, Dreyer, Eisenbraun, Bosse, Wiegand,
  Samek, and Lapuschkin]{achtibat2023attribution}
Reduan Achtibat, Maximilian Dreyer, Ilona Eisenbraun, Sebastian Bosse, Thomas
  Wiegand, Wojciech Samek, and Sebastian Lapuschkin.
\newblock From attribution maps to human-understandable explanations through
  {Concept Relevance Propagation}.
\newblock \emph{Nature Machine Intelligence}, 5\penalty0 (9):\penalty0
  1006--1019, 2023.

\bibitem[Adebayo et~al.(2018)Adebayo, Gilmer, Muelly, Goodfellow, Hardt, and
  Kim]{adebayo2018sanity}
Julius Adebayo, Justin Gilmer, Michael Muelly, Ian Goodfellow, Moritz Hardt,
  and Been Kim.
\newblock Sanity checks for saliency maps.
\newblock In \emph{Proceedings of the 32nd Conference on Advances in Neural
  Information Processing Systems (NeurIPS)}, 2018.

\bibitem[Alvarez-Melis and Jaakkola(2018)]{AlvarezMelis2018towards}
David Alvarez-Melis and Tommi~S. Jaakkola.
\newblock Towards robust interpretability with self-explaining neural networks.
\newblock In \emph{Proceedings of the 32nd Conference on Advances in Neural
  Information Processing Systems (NeurIPS)}, 2018.

\bibitem[Atakishiyev et~al.(2025)Atakishiyev, Salameh, and
  Goebel]{atakishiyev2025safety}
Shahin Atakishiyev, Mohammad Salameh, and Randy Goebel.
\newblock Safety implications of explainable artificial intelligence in
  end-to-end autonomous driving.
\newblock \emph{IEEE Transactions on Intelligent Transportation Systems},
  26\penalty0 (10):\penalty0 14516--14535, 2025.

\bibitem[Bai et~al.(2023)Bai, Yeh, Lin, Ravikumar, and Hsieh]{bai2023concept}
Andrew Bai, Chih-Kuan Yeh, Neil~Y.C. Lin, Pradeep~Kumar Ravikumar, and Cho-Jui
  Hsieh.
\newblock Concept gradient: {Concept-based }interpretation without linear
  assumption.
\newblock In \emph{Proceedings of the 11th International Conference on Learning
  Representations (ICLR)}, 2023.

\bibitem[Bica et~al.(2021)Bica, Jarrett, H{\"u}y{\"u}k, and van~der
  Schaar]{bica2020learning}
Ioana Bica, Daniel Jarrett, Alihan H{\"u}y{\"u}k, and Mihaela van~der Schaar.
\newblock Learning ``what-if'' explanations for sequential decision-making.
\newblock In \emph{Proceedings of the 9th International Conference on Learning
  Representations (ICLR)}, 2021.

\bibitem[Brockman et~al.(2016)Brockman, Cheung, Pettersson, Schneider,
  Schulman, Tang, and Zaremba]{brockman2016openai}
Greg Brockman, Vicki Cheung, Ludwig Pettersson, Jonas Schneider, John Schulman,
  Jie Tang, and Wojciech Zaremba.
\newblock {OpenAI Gym}.
\newblock \emph{arXiv preprint arXiv:1606.01540}, 2016.

\bibitem[Celli(2022)]{celli2022causal}
Viviana Celli.
\newblock Causal mediation analysis in economics: Objectives, assumptions,
  models.
\newblock \emph{Journal of Economic Surveys}, 36\penalty0 (1):\penalty0
  214--234, 2022.

\bibitem[Chen et~al.(2018)Chen, Li, Grosse, and Duvenaud]{chen2018isolating}
Ricky T.~Q. Chen, Xuechen Li, Roger Grosse, and David Duvenaud.
\newblock Isolating sources of disentanglement in variational autoencoders.
\newblock In \emph{Proceedings of the 32nd Conference on Advances in Neural
  Information Processing Systems (NeurIPS)}, 2018.

\bibitem[Das et~al.(2023)Das, Chernova, and Kim]{das2023state2explanation}
Devleena Das, Sonia Chernova, and Been Kim.
\newblock {State2Explanation}: Concept-based explanations to benefit agent
  learning and user understanding.
\newblock In \emph{Proceedings of the 37th Conference on Advances in Neural
  Information Processing Systems (NeurIPS)}, 2023.

\bibitem[De~Felice et~al.(2025)De~Felice, Casanova~Flores, De~Santis, Santini,
  Schneider, Barbiero, and Termine]{de2026causally}
Giovanni De~Felice, Arianna Casanova~Flores, Francesco De~Santis, Silvia
  Santini, Johannes Schneider, Pietro Barbiero, and Alberto Termine.
\newblock Causally reliable concept bottleneck models.
\newblock \emph{Advances in Neural Information Processing Systems},
  38:\penalty0 149099--149139, 2025.

\bibitem[Delfosse et~al.(2024)Delfosse, Sztwiertnia, Rothermel, Stammer, and
  Kersting]{delfosse2024interpretable}
Quentin Delfosse, Sebastian Sztwiertnia, Mark Rothermel, Wolfgang Stammer, and
  Kristian Kersting.
\newblock Interpretable concept bottlenecks to align reinforcement learning
  agents.
\newblock In \emph{Proceedings of the 38th Conference on Advances in Neural
  Information Processing Systems (NeurIPS)}, 2024.

\bibitem[Dominici et~al.(2025{\natexlab{a}})Dominici, Barbiero,
  Espinosa~Zarlenga, Termine, Gjoreski, Marra, and
  Langheinrich]{dominici2025causal}
Gabriele Dominici, Pietro Barbiero, Mateo Espinosa~Zarlenga, Alberto Termine,
  Martin Gjoreski, Giuseppe Marra, and Marc Langheinrich.
\newblock Causal concept graph models: Beyond causal opacity in deep learning.
\newblock In \emph{International Conference on Learning Representations},
  volume 2025, pages 55102--55131, 2025{\natexlab{a}}.

\bibitem[Dominici et~al.(2025{\natexlab{b}})Dominici, Barbiero, Giannini,
  Gjoreski, Marra, and Langheinrich]{dominici2025counterfactual}
Gabriele Dominici, Pietro Barbiero, Francesco Giannini, Martin Gjoreski,
  Giuseppe Marra, and Marc Langheinrich.
\newblock Counterfactual concept bottleneck models.
\newblock In \emph{Proceedings of the 13th International Conference on Learning
  Representations (ICLR)}, 2025{\natexlab{b}}.

\bibitem[Eastwood and Williams(2018)]{eastwood2018framework}
Cian Eastwood and Christopher K.~I. Williams.
\newblock A framework for the quantitative evaluation of disentangled
  representations.
\newblock In \emph{Proceedings of the 6th International Conference on Learning
  Representations (ICLR)}, 2018.

\bibitem[Echterhoff et~al.(2024)Echterhoff, Yan, Han, Abdelraouf, Gupta, and
  McAuley]{echterhoff2024driving}
Jessica Echterhoff, An~Yan, Kyungtae Han, Amr Abdelraouf, Rohit Gupta, and
  Julian McAuley.
\newblock Driving through the concept gridlock: Unraveling explainability
  bottlenecks in automated driving.
\newblock In \emph{Proceedings of the IEEE/CVF Winter Conference on
  Applications of Computer Vision}, pages 7346--7355, 2024.

\bibitem[Geiger et~al.(2021)Geiger, Lu, Icard, and Potts]{geiger2021causal}
Atticus Geiger, Hanson Lu, Thomas Icard, and Christopher Potts.
\newblock Causal abstractions of neural networks.
\newblock \emph{Advances in Neural Information Processing Systems},
  34:\penalty0 9574--9586, 2021.

\bibitem[Geiger et~al.(2022)Geiger, Wu, Lu, Rozner, Kreiss, Icard, Goodman, and
  Potts]{geiger2022inducing}
Atticus Geiger, Zhengxuan Wu, Hanson Lu, Josh Rozner, Elisa Kreiss, Thomas
  Icard, Noah Goodman, and Christopher Potts.
\newblock Inducing causal structure for interpretable neural networks.
\newblock In \emph{International Conference on Machine Learning}, pages
  7324--7338. PMLR, 2022.

\bibitem[Geiger et~al.(2025)Geiger, Ibeling, Zur, Chaudhary, Chauhan, Huang,
  Arora, Wu, Goodman, Potts, and Icard]{geiger2025causal}
Atticus Geiger, Duligur Ibeling, Amir Zur, Maheep Chaudhary, Sonakshi Chauhan,
  Jing Huang, Aryaman Arora, Zhengxuan Wu, Noah Goodman, Christopher Potts, and
  Thomas Icard.
\newblock Causal abstraction: A theoretical foundation for mechanistic
  interpretability.
\newblock \emph{Journal of Machine Learning Research}, 26\penalty0
  (83):\penalty0 1--64, 2025.

\bibitem[Goyal et~al.(2019)Goyal, Feder, Shalit, and Kim]{goyal2019explaining}
Yash Goyal, Amir Feder, Uri Shalit, and Been Kim.
\newblock Explaining classifiers with causal concept effect ({CaCE}).
\newblock \emph{arXiv preprint arXiv:1907.07165}, 2019.

\bibitem[Gyevnar et~al.(2024)Gyevnar, Wang, Lucas, Cohen, and
  Albrecht]{gyevnar2024causal}
Balint Gyevnar, Cheng Wang, Christopher~G. Lucas, Shay~B. Cohen, and Stefano~V.
  Albrecht.
\newblock Causal explanations for sequential decision-making in multi-agent
  systems.
\newblock In \emph{Proceedings of the 23rd International Conference on
  Autonomous Agents and Multiagent Systems (AAMAS)}, 2024.

\bibitem[Heidlauf et~al.(2018)Heidlauf, Collins, Bolender, and
  Bak]{heidlauf2018verification}
Peter Heidlauf, Alexander Collins, Michael Bolender, and Stanley Bak.
\newblock Verification challenges in {F-16} ground collision avoidance and
  other automated maneuvers.
\newblock In \emph{Proceedings of the 5th International Workshop on Applied
  Verification of Continuous and Hybrid Systems (ARCH)}, 2018.

\bibitem[Herlau and Larsen(2022)]{herlau2022reinforcement}
Tue Herlau and Rasmus Larsen.
\newblock Reinforcement learning of causal variables using mediation analysis.
\newblock In \emph{Proceedings of the AAAI Conference on Artificial
  Intelligence}, volume~36, pages 6910--6917, 2022.

\bibitem[Highland et~al.(2023)Highland, Schnell, Woodruff, and
  Avdic-McIntire]{highland2023towards}
Patrick Highland, Thomas Schnell, Katharine Woodruff, and Gianna
  Avdic-McIntire.
\newblock Towards human objective real-time trust of autonomy measures for
  combat aviation.
\newblock \emph{The International Journal of Aerospace Psychology}, 33\penalty0
  (1):\penalty0 1--34, 2023.

\bibitem[Imai et~al.(2010)Imai, Keele, and Tingley]{imai2010general}
Kosuke Imai, Luke Keele, and Dustin Tingley.
\newblock A general approach to causal mediation analysis.
\newblock \emph{Psychological Methods}, 15\penalty0 (4):\penalty0 309--334,
  2010.

\bibitem[Ismail et~al.(2024)Ismail, Adebayo, Corrada~Bravo, Ra, and
  Cho]{ismail2024concept}
Aya~Abdelsalam Ismail, Julius Adebayo, H{\'e}ctor Corrada~Bravo, Stephen Ra,
  and Kyunghyun Cho.
\newblock Concept bottleneck generative models.
\newblock In \emph{International Conference on Learning Representations},
  volume 2024, pages 34114--34132, 2024.

\bibitem[Keki{\'c} et~al.(2026)Keki{\'c}, Schneider, B{\"u}chler,
  Sch{\"o}lkopf, and Besserve]{kekic2025learning}
Armin Keki{\'c}, Jan Schneider, Dieter B{\"u}chler, Bernhard Sch{\"o}lkopf, and
  Michel Besserve.
\newblock Learning nonlinear causal reductions to explain reinforcement
  learning policies.
\newblock In \emph{Proceedings of the 14th International Conference on Learning
  Representations (ICLR)}, 2026.

\bibitem[Kenny et~al.(2023)Kenny, Tucker, and Shah]{kenny2023towards}
Eoin~M Kenny, Mycal Tucker, and Julie Shah.
\newblock Towards interpretable deep reinforcement learning with human-friendly
  prototypes.
\newblock In \emph{Proceedings of the 11th International Conference on Learning
  Representations (ICLR)}, 2023.

\bibitem[Kenny et~al.(2026)Kenny, Dharmavaram, Lee, Phan-Minh, Rajesh, Hu,
  Major, Tomov, and Shah]{kenny2026explainable}
Eoin~M Kenny, Akshay Dharmavaram, Sang~Uk Lee, Tung Phan-Minh, Shreyas Rajesh,
  Yunqing Hu, Laura Major, Momchil~S Tomov, and Julie~A Shah.
\newblock Explainable deep learning improves human mental models of
  self-driving cars.
\newblock \emph{Nature}, 657\penalty0 (8130):\penalty0 114--120, 2026.

\bibitem[Khemakhem et~al.(2020)Khemakhem, Kingma, Monti, and
  Hyv{\"a}rinen]{khemakhem2020variational}
Ilyes Khemakhem, Diederik~P. Kingma, Ricardo~P. Monti, and Aapo Hyv{\"a}rinen.
\newblock Variational autoencoders and nonlinear {ICA}: A unifying framework.
\newblock In \emph{Proceedings of the 23rd International Conference on
  Artificial Intelligence and Statistics (AISTATS)}, 2020.

\bibitem[Kim et~al.(2018)Kim, Wattenberg, Gilmer, Cai, Wexler, Viegas, and
  Sayres]{kim2018interpretability}
Been Kim, Martin Wattenberg, Justin Gilmer, Carrie Cai, James Wexler, Fernanda
  Viegas, and Rory Sayres.
\newblock Interpretability beyond feature attribution: {Quantitative} testing
  with concept activation vectors ({TCAV}).
\newblock In \emph{Proceedings of the 35th International Conference on Machine
  Learning (ICML)}, 2018.

\bibitem[Kim and Mnih(2018)]{kim2018disentangling}
Hyunjik Kim and Andriy Mnih.
\newblock Disentangling by factorising.
\newblock In \emph{Proceedings of the 35th International Conference on Machine
  Learning (ICML)}, 2018.

\bibitem[Koh et~al.(2020)Koh, Nguyen, Tang, Mussmann, Pierson, Kim, and
  Liang]{koh2020concept}
Pang~Wei Koh, Thao Nguyen, Yew~Siang Tang, Stephen Mussmann, Emma Pierson, Been
  Kim, and Percy Liang.
\newblock Concept bottleneck models.
\newblock In \emph{Proceedings of the 37th International Conference on Machine
  Learning (ICML)}, 2020.

\bibitem[Laguna et~al.(2024)Laguna, Marcinkevi{\v{c}}s, Vandenhirtz, and
  Vogt]{laguna2024beyond}
Sonia Laguna, Ri{\v{c}}ards Marcinkevi{\v{c}}s, Moritz Vandenhirtz, and
  Julia~E. Vogt.
\newblock Beyond concept bottleneck models: How to make black boxes
  intervenable?
\newblock \emph{Advances in Neural Information Processing Systems},
  37:\penalty0 85006--85044, 2024.

\bibitem[Lee et~al.(2021)Lee, Cashin, Lamb, Hopewell, Vansteelandt,
  VanderWeele, MacKinnon, Mansell, Collins, Golub, et~al.]{lee2021guideline}
Hopin Lee, Aidan~G Cashin, Sarah~E Lamb, Sally Hopewell, Stijn Vansteelandt,
  Tyler~J VanderWeele, David~P MacKinnon, Gemma Mansell, Gary~S Collins,
  Robert~M Golub, et~al.
\newblock A guideline for reporting mediation analyses of randomized trials and
  observational studies: the {AGReMA} statement.
\newblock \emph{Journal of the American Medical Association}, 326\penalty0
  (11):\penalty0 1045--1056, 2021.

\bibitem[Li et~al.(2025)Li, Zhang, and Bareinboim]{Li2025confounding}
Mingxuan Li, Junzhe Zhang, and Elias Bareinboim.
\newblock Confounding robust deep reinforcement learning: A causal approach.
\newblock In \emph{Proceedings of the 39th Conference on Advances in Neural
  Information Processing Systems (NeurIPS)}, 2025.

\bibitem[Loh et~al.(2022)Loh, Moerkerke, Loeys, and
  Vansteelandt]{loh2022disentangling}
Wen~Wei Loh, Beatrijs Moerkerke, Tom Loeys, and Stijn Vansteelandt.
\newblock Disentangling indirect effects through multiple mediators without
  assuming any causal structure among the mediators.
\newblock \emph{Psychological Methods}, 27\penalty0 (6):\penalty0 982--999,
  2022.

\bibitem[Lundberg and Lee(2017)]{lundberg2017unified}
Scott~M Lundberg and Su-In Lee.
\newblock A unified approach to interpreting model predictions.
\newblock In \emph{Proceedings of the 31st Conference on Advances in Neural
  Information Processing Systems (NeurIPS)}, 2017.

\bibitem[Lyons et~al.(2024)Lyons, Highland, Bos, Lyons, Skinner, Schnell, and
  Hefron]{lyons2024measuring}
Joseph Lyons, Patrick Highland, Nathan Bos, David Lyons, Anna Skinner, Thomas
  Schnell, and Ryan Hefron.
\newblock Measuring perceived agent appropriateness in a live-flight
  human-autonomy teaming scenario.
\newblock \emph{Ergonomics in Design}, 32\penalty0 (4):\penalty0 12--17, 2024.

\bibitem[MacKinnon(2008)]{mackinnon2008introduction}
David~P MacKinnon.
\newblock \emph{Introduction to Statistical Mediation Analysis}.
\newblock Lawrence Erlbaum Associates, 2008.
\newblock ISBN 9780805839746.

\bibitem[Madumal et~al.(2020)Madumal, Miller, Sonenberg, and
  Vetere]{madumal2020explainable}
Prashan Madumal, Tim Miller, Liz Sonenberg, and Frank Vetere.
\newblock Explainable reinforcement learning through a causal lens.
\newblock In \emph{Proceedings of the 34th AAAI Conference on Artificial
  Intelligence (AAAI)}, 2020.

\bibitem[Mahmud et~al.(2024)Mahmud, Saisubramanian, and
  Zilberstein]{mahmud2024verification}
Saaduddin Mahmud, Sandhya Saisubramanian, and Shlomo Zilberstein.
\newblock Verification and validation of {AI} systems using explanations.
\newblock In \emph{Proceedings of the AAAI Symposium on AI Trustworthiness and
  Risk Assessment for Challenging Contexts (ATRACC)}, 2024.

\bibitem[Nashed et~al.(2025)Nashed, Mahmud, Goldman, and
  Zilberstein]{nashed2025causal}
Samer~B Nashed, Saaduddin Mahmud, Claudia~V Goldman, and Shlomo Zilberstein.
\newblock Causal explanations for sequential decision making.
\newblock \emph{Journal of Artificial Intelligence Research}, 83\penalty0 (17),
  2025.

\bibitem[Omeiza et~al.(2022)Omeiza, Webb, Jirotka, and
  Kunze]{omeiza2021explanations}
Daniel Omeiza, Helena Webb, Marina Jirotka, and Lars Kunze.
\newblock Explanations in autonomous driving: A survey.
\newblock \emph{IEEE Transactions on Intelligent Transportation Systems},
  23\penalty0 (8):\penalty0 10142--10162, 2022.

\bibitem[Pearl(2009)]{pearl2009causality}
Judea Pearl.
\newblock \emph{Causality: Models, Reasoning and Inference}.
\newblock Cambridge University Press, USA, 2nd edition, 2009.
\newblock ISBN 052189560X.

\bibitem[Pearl(2012)]{pearl2012causal}
Judea Pearl.
\newblock The causal mediation formula -- {A} guide to the assessment of
  pathways and mechanisms.
\newblock \emph{Prevention Science}, 13\penalty0 (4):\penalty0 426--436, 2012.

\bibitem[Peters et~al.(2017)Peters, Janzing, and
  Sch{\"o}lkopf]{peters2017elements}
Jonas Peters, Dominik Janzing, and Bernhard Sch{\"o}lkopf.
\newblock \emph{Elements of Causal Inference: Foundations and Learning
  Algorithms}.
\newblock The MIT Press, 2017.
\newblock ISBN 0262037319.

\bibitem[Phan-Minh et~al.(2023)Phan-Minh, Howington, Chu, Tomov, Beaudoin, Lee,
  Li, Dicle, Findler, Suarez-Ruiz, Yang, Omari, and Wolff]{phan2022driving}
Tung Phan-Minh, Forbes Howington, Ting-Sheng Chu, Momchil~S Tomov, Robert~E
  Beaudoin, Sang~Uk Lee, Nanxiang Li, Caglayan Dicle, Samuel Findler, Francisco
  Suarez-Ruiz, Bo~Yang, Sammy Omari, and Eric~M Wolff.
\newblock {DriveIRL: D}rive in real life with inverse reinforcement learning.
\newblock In \emph{Proceedings of the 2023 IEEE International Conference on
  Robotics and Automation (ICRA)}, 2023.

\bibitem[Pope et~al.(2023)Pope, Ide, Mi{\'c}ovi{\'c}, Diaz, Twedt, Alcedo,
  Walker, Rosenbluth, Ritholtz, and Javorsek]{pope2022hierarchical}
Adrian~P Pope, Jaime~S Ide, Daria Mi{\'c}ovi{\'c}, Henry Diaz, Jason~C Twedt,
  Kevin Alcedo, Thayne~T Walker, David Rosenbluth, Lee Ritholtz, and Daniel
  Javorsek.
\newblock Hierarchical reinforcement learning for air combat at {DARPA's}
  {AlphaDogfight} trials.
\newblock \emph{IEEE Transactions on Artificial Intelligence}, 4\penalty0
  (6):\penalty0 1371--1385, 2023.

\bibitem[Qian et~al.(2024)Qian, Yu, Wu, Su, Wang, and Fan]{qian2024learning}
Yilue Qian, Peiyu Yu, Ying~Nian Wu, Yao Su, Wei Wang, and Lifeng Fan.
\newblock Learning concept-based causal transition and symbolic reasoning for
  visual planning.
\newblock In \emph{Proceedings of the 2024 IEEE/RSJ International Conference on
  Intelligent Robots and Systems (IROS)}, 2024.

\bibitem[Ribeiro et~al.(2016)Ribeiro, Singh, and Guestrin]{ribeiro2016why}
Marco~Tulio Ribeiro, Sameer Singh, and Carlos Guestrin.
\newblock ``{Why} should {I} trust you?'': Explaining the predictions of any
  classifier.
\newblock In \emph{Proceedings of the 22nd ACM SIGKDD International Conference
  on Knowledge Discovery and Data Mining (KDD)}, 2016.

\bibitem[Rountree et~al.(2021)Rountree, Hipelius, Dienst, Aronoff, Neely,
  Steigerwald, Griffis, de~Schweinitz, Lee, and Hefron]{rountree2021testing}
Joshua Rountree, Patrick Hipelius, Brian Dienst, Jonathan Aronoff, Ryan Neely,
  Robert Steigerwald, Skylar Griffis, David de~Schweinitz, Chiawei Lee, and
  Ryan Hefron.
\newblock Testing artificial intelligence in high-performance, tactical
  aircraft.
\newblock In \emph{Proceedings of the 2021 IEEE Aerospace Conference (50100)},
  2021.

\bibitem[Ruan et~al.(2023)Ruan, Zhang, Di, and Bareinboim]{ruan2023causal}
Kangrui Ruan, Junzhe Zhang, Xuan Di, and Elias Bareinboim.
\newblock Causal imitation learning via inverse reinforcement learning.
\newblock In \emph{Proceedings of the 11th International Conference on Learning
  Representations (ICLR)}, 2023.

\bibitem[Rudin(2019)]{rudin_stop_2019}
Cynthia Rudin.
\newblock Stop explaining black box machine learning models for high stakes
  decisions and use interpretable models instead.
\newblock \emph{Nature Machine Intelligence}, 1\penalty0 (5):\penalty0
  206--215, May 2019.

\bibitem[Sabour et~al.(2017)Sabour, Frosst, and Hinton]{sabour2017dynamic}
Sara Sabour, Nicholas Frosst, and Geoffrey~E Hinton.
\newblock Dynamic routing between capsules.
\newblock In \emph{Proceedings of the 31st Conference on Advances in Neural
  Information Processing Systems (NeurIPS)}, 2017.

\bibitem[Sanneman and Shah(2022)]{sanneman2022situation}
Lindsay Sanneman and Julie~A Shah.
\newblock The situation awareness framework for explainable {AI (SAFE-AI)} and
  human factors considerations for {XAI} systems.
\newblock \emph{International Journal of Human--Computer Interaction},
  38\penalty0 (18-20):\penalty0 1772--1788, 2022.

\bibitem[Sanneman et~al.(2024)Sanneman, Tucker, and
  Shah]{sanneman2024information}
Lindsay Sanneman, Mycal Tucker, and Julie~A Shah.
\newblock An information bottleneck characterization of the
  understanding-workload tradeoff in human-centered explainable {AI}.
\newblock In \emph{Proceedings of the 2024 ACM Conference on Fairness,
  Accountability, and Transparency (FAccT)}, 2024.

\bibitem[Sch{\"o}lkopf et~al.(2021)Sch{\"o}lkopf, Locatello, Bauer, Ke,
  Kalchbrenner, Goyal, and Bengio]{scholkopf2021toward}
Bernhard Sch{\"o}lkopf, Francesco Locatello, Stefan Bauer, Nan~Rosemary Ke, Nal
  Kalchbrenner, Anirudh Goyal, and Yoshua Bengio.
\newblock Toward causal representation learning.
\newblock \emph{Proceedings of the IEEE}, 109\penalty0 (5):\penalty0 612--634,
  2021.

\bibitem[Schulman et~al.(2017)Schulman, Wolski, Dhariwal, Radford, and
  Klimov]{schulman2017proximal}
John Schulman, Filip Wolski, Prafulla Dhariwal, Alec Radford, and Oleg Klimov.
\newblock Proximal policy optimization algorithms.
\newblock \emph{arXiv preprint arXiv:1707.06347}, 2017.

\bibitem[Schulte and Poupart(2025)]{schulte2025when}
Oliver Schulte and Pascal Poupart.
\newblock When should reinforcement learning use causal reasoning?
\newblock \emph{Transactions on Machine Learning Research}, 2025.
\newblock ISSN 2835-8856.

\bibitem[Selvaraju et~al.(2017)Selvaraju, Cogswell, Das, Vedantam, Parikh, and
  Batra]{selvaraju2017grad}
Ramprasaath~R Selvaraju, Michael Cogswell, Abhishek Das, Ramakrishna Vedantam,
  Devi Parikh, and Dhruv Batra.
\newblock Grad-{CAM}: Visual explanations from deep networks via gradient-based
  localization.
\newblock In \emph{Proceedings of the 2017 IEEE International Conference on
  Computer Vision (ICCV)}, 2017.

\bibitem[Simonyan et~al.(2014)Simonyan, Vedaldi, and
  Zisserman]{simonyan2014deep}
Karen Simonyan, Andrea Vedaldi, and Andrew Zisserman.
\newblock Deep inside convolutional networks: Visualising image classification
  models and saliency maps.
\newblock In \emph{ICLR Workshop}, 2014.

\bibitem[Smilkov et~al.(2017)Smilkov, Thorat, Kim, Vi{\'e}gas, and
  Wattenberg]{smilkov2017smoothgrad}
Daniel Smilkov, Nikhil Thorat, Been Kim, Fernanda Vi{\'e}gas, and Martin
  Wattenberg.
\newblock {SmoothGrad}: Removing noise by adding noise.
\newblock In \emph{ICML 2017 Workshop on Visualization for Deep Learning},
  2017.

\bibitem[So and Fan(2023)]{So-RSS-23}
Oswin So and Chuchu Fan.
\newblock Solving stabilize-avoid optimal control via epigraph form and deep
  reinforcement learning.
\newblock In \emph{Proceedings of Robotics: Science and Systems (R:SS)}, 2023.

\bibitem[Sobr{\'\i}n-Hidalgo et~al.(2025)Sobr{\'\i}n-Hidalgo,
  Guerrero-Higueras, and Matell{\'a}n-Olivera]{sobrin2025generating}
David Sobr{\'\i}n-Hidalgo, {\'A}ngel~Manuel Guerrero-Higueras, and Vicente
  Matell{\'a}n-Olivera.
\newblock Generating explanations for autonomous robots: a systematic review.
\newblock \emph{IEEE Access}, 13:\penalty0 20413--20426, 2025.

\bibitem[Sreedharan et~al.(2022)Sreedharan, Kulkarni, and
  Kambhampati]{sreedharan2022explainable}
Sarath Sreedharan, Anagha Kulkarni, and Subbarao Kambhampati.
\newblock \emph{Explainable Human-AI Interaction: A Planning Perspective}.
\newblock Synthesis Lectures on Artificial Intelligence and Machine Learning.
  Morgan \& Claypool Publishers, 2022.
\newblock ISBN 9781636392905.

\bibitem[Sundararajan et~al.(2017)Sundararajan, Taly, and
  Yan]{sundararajan2017axiomatic}
Mukund Sundararajan, Ankur Taly, and Qiqi Yan.
\newblock Axiomatic attribution for deep networks.
\newblock In \emph{Proceedings of the 34th International Conference on Machine
  Learning (ICML)}, 2017.

\bibitem[Tan et~al.(2024)Tan, Zhou, and Chen]{tan2024explain}
Andong Tan, Fengtao Zhou, and Hao Chen.
\newblock Explain via any concept: Concept bottleneck model with open
  vocabulary concepts.
\newblock In \emph{Proceedings of the 18th European Conference on Computer
  Vision (ECCV)}, 2024.

\bibitem[Tang et~al.(2025)Tang, Abbatematteo, Hu, Chandra,
  Mart{\'\i}n-Mart{\'\i}n, and Stone]{tang2025deep}
Chen Tang, Ben Abbatematteo, Jiaheng Hu, Rohan Chandra, Roberto
  Mart{\'\i}n-Mart{\'\i}n, and Peter Stone.
\newblock Deep reinforcement learning for robotics: A survey of real-world
  successes.
\newblock In \emph{Proceedings of the 39th AAAI Conference on Artificial
  Intelligence (AAAI)}, 2025.

\bibitem[{United States Air Force}(2024)]{aetcman11-251-2024}
{United States Air Force}.
\newblock Flying operations: {T-38C} flying fundamentals.
\newblock Manual AETCMAN 11-251, {Department of Defense}, San Antonio, TX,
  September 2024.
\newblock URL
  \url{https://static.e-publishing.af.mil/production/1/aetc/publication/aetcman11-251/aetcman11-251.pdf}.
\newblock 10 September 2024.

\bibitem[{United States Air Force}(2025)]{aetcman11-248-2025}
{United States Air Force}.
\newblock Flying operations: {T-6} primary flying.
\newblock Manual AETCMAN 11-248, {Department of Defense}, San Antonio, TX,
  August 2025.
\newblock URL
  \url{https://static.e-publishing.af.mil/production/1/aetc/publication/aetcman11-248/aetcman11-248.pdf}.
\newblock 13 August 2025.

\bibitem[VanderWeele(2015)]{vanderweele2015explanation}
Tyler~J VanderWeele.
\newblock \emph{Explanation in Causal Inference: Methods for Mediation and
  Interaction}.
\newblock Oxford University Press, 2015.
\newblock ISBN 9780199325870.

\bibitem[Vansteelandt and Daniel(2017)]{vansteelandt2017interventional}
Stijn Vansteelandt and Rhian~M Daniel.
\newblock Interventional effects for mediation analysis with multiple
  mediators.
\newblock \emph{Epidemiology}, 28\penalty0 (2):\penalty0 258--265, 2017.

\bibitem[Verma and Shah(2025)]{verma2025interpretability}
Pulkit Verma and Julie~A Shah.
\newblock Interpretability analysis of symbolic representations for sequential
  decision-making systems.
\newblock In \emph{HRI 2025 Workshop on Explainability for Human-Robot
  Collaboration: Real-World Concerns (X-HRI)}, 2025.

\bibitem[Verma and Srivastava(2024)]{verma2024learning}
Pulkit Verma and Siddharth Srivastava.
\newblock Learning causally accurate models for autonomous assessment of
  deterministic black-box agents.
\newblock Technical Report TR-ASUSCAI-2024-001, Arizona State University, 2024.

\bibitem[Vig et~al.(2020)Vig, Gehrmann, Belinkov, Qian, Nevo, Singer, and
  Shieber]{vig2020investigating}
Jesse Vig, Sebastian Gehrmann, Yonatan Belinkov, Sharon Qian, Daniel Nevo,
  Yaron Singer, and Stuart Shieber.
\newblock Investigating gender bias in language models using causal mediation
  analysis.
\newblock In \emph{Proceedings of the 34th Conference on Advances in Neural
  Information Processing Systems (NeurIPS)}, 2020.

\bibitem[Wah et~al.(2011)Wah, Branson, Welinder, Perona, and
  Belongie]{wah2011caltech}
Catherine Wah, Steve Branson, Peter Welinder, Pietro Perona, and Serge
  Belongie.
\newblock The {Caltech-UCSD Birds-200-2011} dataset.
\newblock Technical Report CNS-TR-2011-001, California Institute of Technology,
  2011.

\bibitem[Ward(2023)]{ward2023us}
Tom Ward.
\newblock The {US Air Force} is moving fast on {AI}-piloted fighter jets.
\newblock \emph{{WIRED}}, March 2023.

\bibitem[Wu et~al.(2023)Wu, D'Oosterlinck, Geiger, Zur, and
  Potts]{wu2023causal}
Zhengxuan Wu, Karel D'Oosterlinck, Atticus Geiger, Amir Zur, and Christopher
  Potts.
\newblock Causal proxy models for concept-based model explanations.
\newblock In \emph{International Conference on Machine Learning}, pages
  37313--37334. PMLR, 2023.

\bibitem[Zhou et~al.(2022)Zhou, Booth, Ribeiro, and Shah]{zhou2022feature}
Yilun Zhou, Serena Booth, Marco~Tulio Ribeiro, and Julie Shah.
\newblock Do feature attribution methods correctly attribute features?
\newblock In \emph{Proceedings of the 36th AAAI Conference on Artificial
  Intelligence (AAAI)}, 2022.

\end{thebibliography}
\normalsize

\appendix

\newpage

\section{Variable dictionary}
\label{sec:variable-dictionary}

{\small
\begin{longtable}{ll}
\caption{Variable dictionary for CCW-Net.}
\label{tab:variables}\\
\toprule
Symbol & Definition \\
\midrule
\endfirsthead
\toprule
Symbol & Definition \\
\midrule
\endhead
\midrule
\endfoot
\bottomrule
\endlastfoot
\multicolumn{2}{l}{\textit{Inputs and outputs}} \\
$X$ & Input \\
$Y$ & Expert output \\
$\hat{Y}$ & Wrapper output, $h_\phi(M)$ \\
$c_j \in \{0, 1\}$ & Concept label ($1 =$ active) \\
\midrule
\multicolumn{2}{l}{\textit{Architecture}} \\
$f$ & Frozen backbone \\
$Z = f(X)$ & Backbone latent \\
$g_\theta$ & Concept encoder \\
$h_\phi$ & Attention head \\
$M = (M_1, \dots, M_J)$ & Concept representation \\
$M_j \in \mathbb{R}^D$ & Concept $j$'s representation (slot) \\
$J$ & Number of concepts \\
$D$ & Representation dimension \\
\midrule
\multicolumn{2}{l}{\textit{Mediator model (Sec.~\ref{sec:method-flow})}} \\
$\mathcal{F}_\psi$ & Per-concept invertible flow \\
$\tilde{M}_j$ & Counterfactual draw, $\tilde{M}_j \sim \mathcal{F}_\psi(\cdot \mid 1 - c_j, X)$ \\
$\mathcal{L}_{\text{flow}}$ & Flow negative log-likelihood \\
\midrule
\multicolumn{2}{l}{\textit{Mediation analysis (Sec.~\ref{sec:problem})}} \\
$\mathrm{TE}, \mathrm{DE}, \mathrm{IE}$ & Total, direct, indirect effect \\
$\mathrm{IE}_j$ & Interventional indirect effect of mediator $j$ \\
$\mathrm{IE}_\mu$ & Residual mediator interaction \\
\midrule
\multicolumn{2}{l}{\textit{Concept edits and changes (Sec.~\ref{sec:method-criteria})}} \\
$\mathbf{0}$ & All-zero concept representation \\
$\mathbf{0}_{j \leftarrow M_j}$ & $\mathbf{0}$ with slot $j$ set to $M_j$ \\
$M_{j \leftarrow \tilde{M}_j}$ & $M$ with slot $j$ set to $\tilde{M}_j$ \\
$\Delta^0$ & Change from zero, $\hat{Y} - h_\phi(\mathbf{0})$ \\
$\Delta^0_j$ & Slot $j$'s change from zero, $h_\phi(\mathbf{0}_{j \leftarrow M_j}) - h_\phi(\mathbf{0})$ \\
$\Delta^{\text{cf}}_j$ & Slot $j$'s counterfactual change, $h_\phi(M_{j \leftarrow \tilde{M}_j}) - \hat{Y}$ \\
\midrule
\multicolumn{2}{l}{\textit{Losses and weights}} \\
$\mathcal{L}_{\text{task}}$ & Task loss \\
$\mathcal{L}_{\text{cls}}$ & Concept classification loss \\
$\mathcal{L}_{\text{nec}}, \mathcal{L}_{\text{suf}}, \mathcal{L}_{\text{ind}}$ & Necessity, sufficiency, independence losses \\
$\mathcal{L}_{\text{pres}}$ & Presence loss (auxiliary) \\
$\mathcal{L}_{\text{align}}$ & Weighted sum of the alignment losses \\
$\mathcal{L}_{\text{wrap}}$ & Total wrapper loss \\
$\lambda_{\text{cls}}, \lambda_{\text{nec}}, \dots, \lambda_{\text{pres}}$ & Loss weights \\
$\beta$ & Presence threshold, $1/\sum_j c_j$ \\
\midrule
\multicolumn{2}{l}{\textit{Evaluation (Sec.~\ref{sec:experiments-eval})}} \\
$\mathcal{A}, \mathcal{B}$ & Test-input sets of one comparison \\
$z_i$ & Matching features of input $i$ \\
$d_{ii'}$ & Distance between $z_i$ and $z_{i'}$ \\
$\tau$ & Kernel bandwidth \\
$G$, $\pi$ & Slot group; group ordering \\
$M^{\mathcal{A}}_G, M^{\mathcal{B}}_G$ & Group $G$'s slots from an $\mathcal{A}$ or $\mathcal{B}$ match \\
$\mathrm{IE}_G$ & Group $G$'s effect \\
$\rho_j$ & Fraction of the effect slot $j$ carries \\
$\rho_{\text{active}}, \rho_{\text{inactive}}$ & Fraction active, inactive concepts carry \\
\midrule
\multicolumn{2}{l}{\textit{Evaluation estimator (App.~\ref{sec:estimator-details})}} \\
$\ell$, $n_\ell$ & Concept label combination, its test count \\
$c$, $\mathcal{C}(\ell)$ & Comparison, comparisons that include label combination $\ell$ \\
$G^{\mathrm{sw}}_c$ & Swapped pair of comparison $c$ \\
$\mathcal{S}$ & One of the sets $\mathcal{A}, \mathcal{B}$ \\
$i'$ & Match of test input $i$ \\
$K$ & Number of matching features \\
$\epsilon$ & Matching radius \\
$\mathrm{IE}^{(c)}_G$ & Group $G$'s effect in comparison $c$ \\
$\mathcal{C}_j(\ell)$ & Comparisons in $\mathcal{C}(\ell)$ with $j \notin G^{\mathrm{sw}}_c$ \\
$G_c(j)$ & Concept slot $j$'s group in comparison $c$ \\
$\|\mathrm{IE}_j\|^{(\ell)}$, $\rho^{(\ell)}_j$ & Concept slot $j$'s effect, fraction for label combination $\ell$ \\
$\Lambda$ & Scored combinations with an active concept \\
$\bar{\rho}_{\text{active}}, \bar{\rho}_{\text{inactive}}$ & Test-count-weighted means over $\Lambda$ \\
\end{longtable}}

\section{Extended related work}
\label{sec:extended-related-work}

In Section~\ref{sec:related-work} we focused on the most directly relevant prior art for CCW-Net's three contributions (a mediation-analysis training objective, an architecture for concept-level mediation analysis, and an evaluation method for concept-based causality claims). Here we position CCW-Net among causal concept bottleneck models and causal abstraction methods, and discuss adjacent research operating at different levels (input, environment-state, policy, trajectory) or in adjacent approaches.

\subsection{Causal concept bottleneck models and causal abstraction}

Concept bottleneck models differ in how concepts reach and produce the model's output. In flat bottlenecks, every concept is a direct parent of the output. \citet{laguna2024beyond} make pretrained black boxes intervenable, fine-tuning them such that concept interventions improve their predictions. \citet{ismail2024concept} insert a concept bottleneck into generative models to steer generation toward human-interpretable concepts. \citet{dominici2025counterfactual} extend concept bottleneck models to predict how concepts would need to change to produce an alternative prediction. These works use concept interventions to improve or steer predictions. CCW-Net instead trains and measures how much of a model's output is attributed to each concept. Concept graph models \citep{de2026causally,dominici2025causal} let some concepts be parents of others rather than of the output, allowing for a deeper, more intricate concept graph. Driving and measuring the causal relationship of each node in such a graph is a natural extension of our approach. 

In the causal abstraction literature, interchange intervention training \citep{geiger2022inducing} trains a network so that interventions on aligned high-level variables produce the counterfactual outputs of a manually-specified causal model. Causal proxy models \citep{wu2023causal} train a proxy of a text classifier on counterfactual examples so that concept interventions simulate the classifier's counterfactual behavior. CCW-Net needs neither a hand-specified causal model nor counterfactual input pairs. Its losses follow interventional mediation analysis \citep{vansteelandt2017interventional}, which also requires no causal ordering among the mediators. For measurement of causal effects, the causal concept effect (CaCE) of \cite{goyal2019explaining} estimates the change in expected output when a concept is toggled, using a learned generative model of counterfactual inputs and averaging over all samples. CCW-Net trains for the property and measures it separately for active and inactive concepts. 

\subsection{Causal explanation in sequential decision-making}
Several lines of work explain sequential decision-making policy behavior using causal frameworks at the environment-state, trajectory, or action level. \citet{madumal2020explainable} introduce action-influence models that define a structural causal model over environment variables and agent actions and use them to produce contrastive explanations of RL policies. \citet{kekic2025learning} learn nonlinear causal reductions that map a high-dimensional RL system to a lower-dimensional SCM whose interventional behavior matches the original system. \citet{nashed2025causal} build a unified SCM framework for explaining the actions of MDP agents, and \citet{gyevnar2024causal} extend causal explanations to multi-agent settings. \citet{bica2020learning} explain expert sequential decision-making policies by learning rewards over what-if outcomes. \citet{schulte2025when} examine when reinforcement learning benefits from causal reasoning. \citet{verma2024learning} learn causally accurate models for autonomous assessment of black-box agents. These methods operate at the environment or policy level and do not impose causal structure on a learned concept representation. CCW-Net is complementary: it uses concept labels to shape the model's own concept-to-output effects, without requiring an explicit environment SCM.

\subsection{Causal methods in imitation and reinforcement learning training}

Causal frameworks have also been incorporated into imitation learning and reinforcement learning training, distinct from post-hoc explanation. \citet{ruan2023causal} introduce causal imitation learning via inverse reinforcement learning to recover policies that respect causal structure in expert demonstrations. \citet{Li2025confounding} address confounding-robust deep RL via a causal approach to off-policy learning. \citet{herlau2022reinforcement} learn causal variables and a policy by optimizing a mediation-analysis objective in which the treatment is the choice of policy, the mediator is a binary environment event, and the outcome is the return. These works apply causal reasoning at the policy or imitation level. CCW-Net applies causal reasoning at the concept level inside the policy, training the concept encoder to satisfy interventional necessity, sufficiency, and independence criteria.

\subsection{Human-AI interaction, trust, and verification}

Several lines of work motivate interpretable models in safety-critical settings from human-factors and verification perspectives. \citet{sanneman2022situation,sanneman2024information} characterize the role of explanations in supporting human situational awareness and the understanding-workload tradeoff in human-centered XAI. \citet{sreedharan2022explainable} provide a planning perspective on explainable human-AI interaction. \citet{sobrin2025generating} survey explanation generation for autonomous robots. \citet{lyons2024measuring,highland2023towards} report empirical studies of trust and agent appropriateness in live-flight human-autonomy teaming. \citet{omeiza2021explanations} survey explanation methods specific to autonomous driving. \citet{verma2025interpretability,mahmud2024verification} consider verification and interpretability analysis of decision-making systems. CCW-Net sits within this broader effort: it produces explanations grounded in human-meaningful concepts and aligned with causal effects, supporting the verification, audit, and trust workflows these works study.

\section{Connection to interventional mediation analysis}
\label{sec:vd-connection}

In this section, we connect CCW-Net's method and alignment losses (Sec.~\ref{sec:method-criteria}) to interventional mediation analysis \citep{vansteelandt2017interventional,loh2022disentangling}, and state the attention head property that makes $M_j=0$ a valid reference (App.~\ref{sec:per-slot-identifiability}).

\paragraph{Interventional mediation analysis.}
Mediation analysis asks how much of a treatment's effect on an outcome passes through each of several intermediate variables (i.e., the mediators). Following the notation of \citet{vansteelandt2017interventional}, we write $Y$ for the outcome and $M = (M_1, \dots, M_J)$ for the mediators. $M(a)$ denotes the mediator values under treatment $a$, and $Y(a, m)$ denotes the outcome under treatment $a$ with the mediators set to $m$. Changing the treatment from $a$ to $a^*$ has the total effect
\begin{equation}
\mathrm{TE} = \mathbb{E}\big[Y(a^*, M(a^*)) - Y(a, M(a))\big].
\end{equation}
\citet{vansteelandt2017interventional} measure the part of this effect that passes through a single mediator $M_j$ with the interventional indirect effect
\begin{equation}
\mathrm{IE}_j = \mathbb{E}\big[Y(a^*, M_j(a^*), M_{-j}(a)) - Y(a^*, M_j(a), M_{-j}(a))\big].
\label{eq:vd-iej}
\end{equation}
$\mathrm{IE}_j$ sets only $M_j$ to its value under the new treatment and keeps every other mediator at its value under the original treatment. The direct effect $\mathrm{DE} = \mathbb{E}[Y(a^*, M(a)) - Y(a, M(a))]$ is the part of the effect that bypasses the mediators. \citet{vansteelandt2017interventional} and \citet{loh2022disentangling} add a residual term $\mathrm{IE}_\mu$ for the interactions among mediators. With it, the effects add up to the total without assuming any causal ordering or independence among the mediators,
\begin{equation}
\mathrm{TE} = \mathrm{DE} + \sum_{j=1}^{J} \mathrm{IE}_j + \mathrm{IE}_\mu.
\label{eq:vd-loh-decomp}
\end{equation}

\paragraph{Mapping to CCW-Net.}
In CCW-Net, the concept representations $M_j$ are the mediators and the head output $\hat{Y}$ is the outcome. As the concept representations serve as a bottleneck, the input reaches $\hat{Y}$ only through them with no direct pathway from input to output that bypasses the mediators, so $\mathrm{DE} = 0$ (Sec.~\ref{sec:method-arch}). Flipping concept $j$'s label plays the role of the treatment change, and the flow turns this flip into the counterfactual draw $\tilde{M}_j$ (Eq.~\ref{eq: counterfactual draw}). In the notation above, the treatment value is concept $j$'s label, $a = c_j$, and $\tilde{M}_j$ plays the role of $M_j(a^*)$ with $a^* = 1 - c_j$. With this mapping, two of the changes in Sec.~\ref{sec:method-criteria} correspond to $\mathrm{IE}_j$.
\begin{itemize}[leftmargin=1.3em]
  \item $\Delta^{\text{cf}}_j$ is the per-sample analog of $\mathrm{IE}_j$. Like $\mathrm{IE}_j$, it replaces $M_j$ with its counterfactual value and keeps the other concept representations at their observed values.
  \item $\Delta^0_j$ is the analog of $\mathrm{IE}_j$ measured from zero. It switches $M_j$ on with every other concept representation at zero instead of at its observed value. Zero is a valid reference because $h_\phi(\mathbf{0})$ is a constant (Prop.~\ref{prop:per-slot}).
\end{itemize}

\paragraph{What each loss enforces.}
\begin{itemize}[leftmargin=1.3em]
  \item \textit{Necessity.} $\mathcal{L}_{\text{nec}}$ drives $\Delta^{\text{cf}}_j$ toward zero for concepts labeled inactive, so an inactive concept has no per-sample indirect effect on the output.
  \item \textit{Sufficiency.} $\mathcal{L}_{\text{suf}}$ anchors $h_\phi(\mathbf{0}) + \sum_{j:\, c_j = 1} \Delta^0_j$ to the expert output $Y$. Since $\mathrm{DE} = 0$, the active concepts' contributions measured from zero should reproduce the expert output.
  \item \textit{Independence.} $\mathcal{L}_{\text{ind}}$ drives $\Delta^0 - \sum_j \Delta^0_j$ toward zero. This residual plays the role of $\mathrm{IE}_\mu$ in Eq.~\eqref{eq:vd-loh-decomp} for changes measured from zero, so driving it to zero makes the concept contributions add up.
\end{itemize}
$\mathcal{L}_{\text{pres}}$ is not part of the decomposition. It keeps a few active concepts from carrying the whole effect. Without it, sufficiency could be met overall while the individual contributions say little.

\subsection{Zero value contribution of a zeroed concept}
\label{sec:per-slot-identifiability}

Zero can only serve as a reference if a zeroed concept representation contributes nothing to the output. Here we show that the attention head has this property, and we describe what zeroing a concept does to the other concepts' attention weights.

\paragraph{Setup.}
The wrapper output $\hat{Y}$ is a vector indexed by $r$. The head is a single-head attention layer with one learned query $q_r$ per output. $W_Q$, $W_K$, $W_V$, and $W_O$ are the head's query, key, value, and output projections, all of which are bias-free, and $w_r$ and $b_r$ are the weight and bias for output $r$. We use $j$, $k$, and $l$ to index the $J$ concept representations, where $j$ is the one intervened on, $k$ is any other, and $l$ is a summation index. For output $r$, the head computes logits $s_{rj} = (W_Q q_r)^\top W_K M_j / \sqrt{D}$, weights $\alpha_{rj} = e^{s_{rj}} / \sum_l e^{s_{rl}}$, context $c_r = \sum_l \alpha_{rl} W_O W_V M_l$, and output $\hat{Y}_r = \tanh(w_r^\top c_r + b_r)$.

\begin{definition}[Zero value contribution]
Concept $j$'s value contribution to output $r$ is $\kappa_{rj}(M) = \alpha_{rj}(M)\, W_O W_V M_j$. The head has zero value contribution if $\kappa_{rj}(M) = 0$ for every concept $j$, every output $r$, and every $M$ with $M_j = 0$.
\end{definition}

\begin{proposition}
\label{prop:per-slot}
(i) The head has zero value contribution, and its output at $M = \mathbf{0}$ is the constant $\hat{Y}_r = \tanh(b_r)$. (ii) A concept with $M_j = 0$ keeps softmax weight $\alpha_{rj} = 1/Z_r > 0$ with $Z_r = \sum_{l \neq j} e^{s_{rl}} + 1$, so zeroing a concept is not the same as deleting it. (iii) Any single-concept intervention $M_j \to M_j'$ rescales the other concepts' weights by one common factor per output, $\alpha'_{rk} = \gamma_r \alpha_{rk}$ for all $k \neq j$, with
\begin{equation*}
\gamma_r = \frac{\sum_{l \neq j} e^{s_{rl}} + e^{s_{rj}}}{\sum_{l \neq j} e^{s_{rl}} + e^{s'_{rj}}}.
\end{equation*}
For the zero intervention $M_j' = 0$, the denominator equals the $Z_r$ of (ii), and the effect on the context is exactly $c_r - c'_r = \kappa_{rj} + (1 - \gamma_r)(c_r - \kappa_{rj})$.
\end{proposition}

\begin{proof}
(i) The projections are linear, so $M_j = 0$ gives $W_V M_j = 0$ and hence $\kappa_{rj} = \alpha_{rj} W_O W_V M_j = 0$ for every $r$, whatever the value of $\alpha_{rj}$. At $M = \mathbf{0}$, every logit is $0$, the weights are uniform, and $c_r = 0$, so $\hat{Y}_r = \tanh(b_r)$. (ii) A zero key gives a zero logit rather than a zero weight, since $e^0 = 1$. Deleting the concept would instead give the denominator $Z_r - 1$. (iii) Each logit depends only on its own concept, so the intervention changes only $s_{rj}$. For $k \neq j$, the ratio $\alpha'_{rk} / \alpha_{rk}$ is the ratio of the two denominators, which does not depend on $k$. The zeroed concept's own term vanishes by (i), so $c'_r = \gamma_r (c_r - \kappa_{rj})$, which gives the identity.
\end{proof}

In words, zeroing a concept removes its own value term exactly. Its attention weight stays positive, and the other concepts' weights all scale by the same factor $\gamma_r$.

\paragraph{Relation to ignorability.}
Prop.~\ref{prop:per-slot} complements the ignorability condition in App.~\ref{sec:causal-assumptions}. Ignorability (Prop.~\ref{prop:ignorability}) handles confounding between the mediators and the outcome, $\hat{Y}(m) \perp M \mid X$. Prop.~\ref{prop:per-slot} gives every intervention a zero reference that itself contributes no value to the output.

\section{Causal assumptions}
\label{sec:causal-assumptions}

\begin{figure}[h]
\centering
\begin{tikzpicture}[
 var/.style={draw, rounded corners, align=center, minimum width=10mm,
 minimum height=6mm, fill=gray!10, font=\small},
 arrow/.style={-{Latex[length=2mm]}, thick}
]

\node[var] (X) at (0,0) {$X$};
\node[var] (Z) at (1.8,0) {$Z$};
\node[var] (M1) at (3.6,1.0) {$M_1$};
\node[var] (M2) at (3.6,0.3) {$M_2$};
\node[font=\small] at (3.6,-0.4) {$\vdots$};
\node[var] (MJ) at (3.6,-1.1) {$M_J$};
\node[var] (Y) at (5.2,0) {$\hat Y$};

\draw[arrow] (X) -- node[midway, above, font=\scriptsize] {$f$} (Z);
\draw[arrow] (Z) -- (M1);
\draw[arrow] (Z) -- (M2);
\draw[arrow] (Z) -- (MJ);
\node[font=\scriptsize] at (2.7,1.05) {$g_\theta$};
\draw[arrow] (M1) -- (Y);
\draw[arrow] (M2) -- (Y);
\draw[arrow] (MJ) -- (Y);
\node[font=\scriptsize] at (4.5,1.05) {$h_\phi$};

\end{tikzpicture}

\caption{The CCW-Net wrapper SCM. Every path from $X$ to $\hat{Y}$ passes through the concept representation $M$, and there is no direct $Z \to \hat{Y}$ edge. This gives mediator-outcome ignorability within the SCM (Prop.~\ref{prop:ignorability}).}
\label{fig:wrapper-scm}
\end{figure}
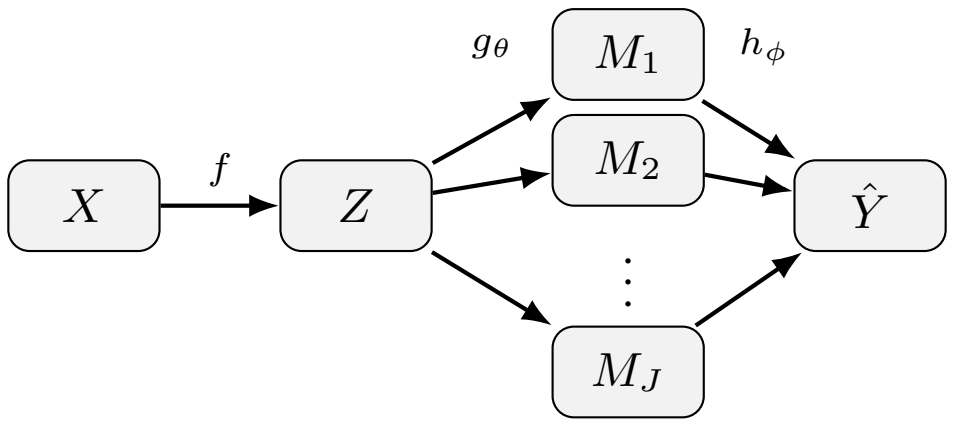

\subsection{Identification assumptions for interventional effects}
\label{sec:identification}

Both training and evaluation estimate interventional effects in the wrapper SCM
\[
X \to Z \to M \to \hat{Y},
\]
where $Z = f(X)$, $M = g_\theta(Z)$, and $\hat{Y} = h_\phi(M)$ (Fig.~\ref{fig:wrapper-scm}). Following \citet{vansteelandt2017interventional} and \citet{loh2022disentangling}, these effects are identified under the following standard conditions.

\begin{itemize}[leftmargin=1.3em]
\item \textbf{Ignorability.} There is no unmeasured confounding of the mediator-outcome relation given $X$, that is, $\hat{Y}(m) \perp M \mid X$ in the wrapper SCM. Prop.~\ref{prop:ignorability} shows that this holds by construction.

\item \textbf{Positivity.} For every $x$ in the support of $X$, the conditional distribution $p(M_j \mid X = x)$ has support around the observed concept representations. This condition concerns the counterfactual draws that the mediator model supplies to the necessity loss during training. The zero reference used by the sufficiency and independence losses is instead defined by the architecture (Prop.~\ref{prop:per-slot}), not a property that occurs in the data.

\item \textbf{Overlap.} In evaluation (Sec.~\ref{sec:experiments-eval}), the mediator model is replaced with real data test inputs. For each test input $i$, both sets $\mathcal{A}$ and $\mathcal{B}$ should contain inputs within a radius of $i$, so that the matched concept representations come from comparable state configurations. When no input is within the radius, the estimator uses the nearest neighbor.

\item \textbf{Consistency.} For the realized concept representation $M = m$, the observed $\hat{Y}$ equals $\hat{Y}(m)$ under the head $h_\phi$.
\end{itemize}

Under these conditions, the effects decompose as in Eq.~\eqref{eq:vd-loh-decomp}, with $\mathrm{IE}_\mu$ collecting the interactions that cannot be attributed to any single concept \citep{loh2022disentangling}.

\subsection{Model-level ignorability in the wrapper SCM}
\label{sec:model-ignorability}

The ignorability condition above holds by construction in the wrapper SCM. The following proposition shows this holds even when the wrapper has stochastic exogenous inputs such as in stochastic policies or with dropout. 

\begin{proposition}[Model-level mediator-outcome ignorability]
\label{prop:ignorability}
Consider the wrapper SCM with structural equations
\[
Z = f(X, U_f), \quad M = g_\theta(Z, U_g), \quad \hat{Y} = h_\phi(M, U_h),
\]
where $U_f, U_g, U_h$ are exogenous noise variables. Suppose:
\begin{enumerate}[leftmargin=1.4em,itemsep=2pt]
\item $U_h \perp (X, U_f, U_g)$, so the head's noise is independent of everything upstream.
\item All randomness used for counterfactual sampling from the mediator model $\mathcal{F}_\psi$ depends only on $(X, U_s)$ with $U_s \perp U_h$.
\end{enumerate}
Then for every concept representation value $m$ in the support of $M \mid X$,
\[
\hat{Y}(m) \perp M \mid X.
\]
Hence the mediator-outcome ignorability required to identify $\mathrm{IE}_j$, $\mathrm{TE}$, and $\mathrm{IE}_\mu$ holds within the wrapper SCM.
\end{proposition}

\begin{proof}
For a fixed $m$, the potential outcome $\hat{Y}(m) = h_\phi(m, U_h)$ is a function of $U_h$ alone. $M$ is a function of $(X, U_f, U_g)$, and $U_h$ is independent of $(X, U_f, U_g)$ by assumption 1. Hence $\hat{Y}(m)$ is independent of $(X, M)$, and in particular $\hat{Y}(m) \perp M \mid X$. A counterfactual draw is a function of $(X, U_s)$, and $U_s \perp U_h$ by assumption 2, so the same argument applies with the draw in place of $M$.
\end{proof}

\section{Architecture details}
\label{sec:architecture-details}

\subsection{Concept encoder $g_\theta$}

The concept encoder is a three-layer multi-layer perceptron with ReLU activations. It maps the backbone latent $Z = f(X)$ to $J$ concept representations,
\begin{equation}
g_\theta(Z) = \mathrm{MLP}(Z) \in \mathbb{R}^{J \times D}.
\end{equation}
The hidden width is 64 for the models in Tables~\ref{tab:aircraft-per-state} and~\ref{tab:carracing-per-state}, and 256 for the CUB-200 models in Table~\ref{tab:cub-per-bucket} and the ablations in App.~\ref{sec:appendix-ablations} outside its last group. We use $D = 16$ on aircraft control and $D = 1$ on CarRacing and CUB-200, and the ablations in App.~\ref{sec:appendix-ablations} vary $D$.

Each concept representation then passes through the capsule-style squash nonlinearity \citep{sabour2017dynamic},
\begin{equation}
\mathrm{squash}(v) = \frac{\|v\|^2}{1 + \|v\|^2} \cdot \frac{v}{\|v\|},
\end{equation}
applied to each of the $J$ concepts separately. The squash bounds each representation's length to $[0, 1)$, and the concept classification loss trains this length to cluster near $0$ and $1$ (avoiding the mid ranges) as the concept's presence (Sec.~\ref{sec:method-arch}).

\subsection{Attention head $h_\phi$}

The head is a single-head attention layer with one learned query $q_r \in \mathbb{R}^D$ per output $r$. The concept representations $M \in \mathbb{R}^{J \times D}$ serve as both keys and values. The query, key, value, and output projections $W_Q, W_K, W_V, W_O \in \mathbb{R}^{D \times D}$ are all bias-free. For output $r$, the head computes the context
\begin{equation}
c_r = \sum_{j=1}^{J} \alpha_{rj}\, W_O W_V M_j, \qquad \alpha_{rj} = \mathrm{softmax}_j\!\left( \frac{(W_Q q_r)^\top W_K M_j}{\sqrt{D}} \right),
\end{equation}
and the output
\begin{equation}
\hat{Y}_r = \tanh\!\left(w_r^\top c_r + b_r\right),
\end{equation}
where $w_r \in \mathbb{R}^D$ and $b_r \in \mathbb{R}$ are the weight and bias for output $r$. Since the projections are bias-free, a zero concept representation produces a zero key and a zero value. It therefore contributes no value to $c_r$, even though its attention weight stays positive (Prop.~\ref{prop:per-slot}). On CUB-200, the tanh outputs of CCW-Net and the linear outputs of CBM are trained with the mean-squared error to the one-hot species label.

\subsection{Per-concept invertible flow $\mathcal{F}_\psi$}

Our chosen mediator model is a conditional normalizing flow with one independent flow per concept. Any modeling approach may be used. Each flow is conditioned on the labels of concept $j$'s mutually exclusive set, the input, and a learned vector identifying concept $j$. For $D > 1$, each flow is a stack of 8 affine coupling layers whose conditioner networks have hidden width 256. For $D = 1$, a coupling layer cannot split the representation, so each flow is instead a stack of 8 conditional affine maps with hidden width 256. To draw a counterfactual for concept $j$, we flip the labels of its set and sample from the flow conditioned on them and the input (Eq.~\ref{eq: counterfactual draw}). The flow has its own optimizer (App.~\ref{sec:training-details}) and is trained on the negative log-likelihood of the encoder's concept representations.

\section{Evaluation estimator}
\label{sec:estimator-details}

In short, we compare similar test inputs that differ in one concept and measure how much of the change in the model's output passes through each concept's representation (i.e., the fraction of change). The causal attribution estimator (Sec.~\ref{sec:experiments-eval}) computes each reported fraction as follows. Let $\ell$ index concept label combinations, $n_\ell$ be the number of test inputs with concept label combination $\ell$, and $\mathcal{C}(\ell)$ be the comparisons in which $\ell$ forms one of the two sets $\mathcal{A}$ and $\mathcal{B}$ (test inputs with original and swapped concept labels). Each comparison $c$ swaps one active concept for a mutually exclusive alternative, so exactly two concept slots change labels. These two concept slots form one group, the swapped pair $G^{\mathrm{sw}}_c$, and every other concept slot forms its own group.

\paragraph{Matching.} We z-score the matching features $z \in \mathbb{R}^K$ (Sec.~\ref{sec:experiments-eval}) over the test split. For a test input $i$ and a set $\mathcal{S} \in \{\mathcal{A}, \mathcal{B}\}$, a match $i' \in \mathcal{S}$ is drawn with probability
\begin{equation}
p(i' \mid i) \propto \exp\!\big(-d_{ii'}^2 / 2\tau^2\big)\, \mathbb{1}[d_{ii'} \le \epsilon]\, \mathbb{1}[i' \neq i],
\qquad
d_{ii'} = \Big(\frac{1}{K} \sum_{k=1}^{K} (z_{ik} - z_{i'k})^2\Big)^{1/2},
\end{equation}
with bandwidth $\tau = 0.25$ and radius $\epsilon = 1.5$. When no input of $\mathcal{S}$ is within the radius, we take the nearest one. Each input receives 96 independently drawn matches for each group and set.

\paragraph{Comparisons.} We form a comparison only when $|\mathcal{A}| \ge 100$ and $|\mathcal{B}| \ge 100$. For each concept label combination $\ell$, we fix one random subset of $\min(n_\ell, 1000)$ test inputs, and these inputs serve as $i$ in every comparison in $\mathcal{C}(\ell)$. Matches come from all test inputs of each set.

\paragraph{Group effects.} $\mathrm{IE}^{(c)}_G$ is Eq.~\eqref{eq:ifm} for comparison $c$, averaged over the inputs, the drawn matches, and all orderings of the groups. We use the Euclidean norm $\|\mathrm{IE}^{(c)}_G\|$ of this averaged change in the model's output.

\paragraph{Concept slot effects.} For concept label combination $\ell$, let $\mathcal{C}_j(\ell) = \{c \in \mathcal{C}(\ell) : j \notin G^{\mathrm{sw}}_c\}$ be the comparisons that keep concept slot $j$ outside the swapped pair, and let $G_c(j)$ be concept slot $j$'s group in comparison $c$. Concept slot $j$'s effect is
\begin{equation}
\|\mathrm{IE}_j\|^{(\ell)} =
\begin{cases}
\dfrac{1}{|\mathcal{C}_j(\ell)|} \displaystyle\sum_{c \in \mathcal{C}_j(\ell)} \big\|\mathrm{IE}^{(c)}_{G_c(j)}\big\| & \text{if } \mathcal{C}_j(\ell) \neq \emptyset, \\[2ex]
\dfrac{1}{|\mathcal{C}(\ell)|} \displaystyle\sum_{c \in \mathcal{C}(\ell)} \big\|\mathrm{IE}^{(c)}_{G^{\mathrm{sw}}_c}\big\| & \text{if } \mathcal{C}_j(\ell) = \emptyset \text{ and } c_j = 1.
\end{cases}
\end{equation}
In the second case, concept slot $j$ is active and lies in the swapped pair of every comparison in $\mathcal{C}(\ell)$. In any other case, concept label combination $\ell$ is not scored. The fractions $\rho^{(\ell)}_j$, $\rho^{(\ell)}_{\text{active}}$, and $\rho^{(\ell)}_{\text{inactive}}$ then follow from Eq.~\eqref{eq:rho}.

\paragraph{Overall values.} All Active and All Inactive weight the scored concept label combinations $\Lambda$ that have an active concept by their test counts,
\begin{equation}
\bar{\rho}_{\text{active}} = \frac{\sum_{\ell \in \Lambda} n_\ell\, \rho^{(\ell)}_{\text{active}}}{\sum_{\ell \in \Lambda} n_\ell},
\qquad
\bar{\rho}_{\text{inactive}} = 1 - \bar{\rho}_{\text{active}},
\end{equation}
and we report their means over ten seeds with two-sided 95\% Student-$t$ intervals.

\paragraph{Effect decomposition.} The total effect, measured on the real matched inputs, equals the sum of the group effects plus a residual $\mathrm{IE}_\mu$ (Eq.~\eqref{eq:vd-loh-decomp}), and this identity holds exactly. Here $\mathrm{IE}_\mu$ is the difference between real inputs and hybrids whose groups come from independent matches.

\section{Training details}
\label{sec:training-details}

Algorithm~\ref{alg:ccw-net} outlines CCW-Net training. The wrapper $(g_\theta, h_\phi)$ and the flow $\mathcal{F}_\psi$ have separate optimizers. Training is conducted in three phases. During warmup, the wrapper learns the task and concept classification. During the intermission, the mediator model is fit to the encoder's concept representations. During alignment, the wrapper trains on the full objective (Sec.~\ref{sec:training objective}), while the mediator model supplies the counterfactual draws for the necessity loss. The mediator model is refit every $E_r$ epochs during alignment so that it remains current with the encoder. Since the CUB-200 models train without the alignment losses they remain in the warmup phase throughout.

\begin{algorithm}
\caption{CCW-Net training. $p_\psi$ denotes the flow's conditional density.}
\label{alg:ccw-net}
\begin{algorithmic}[1]
\Require Dataset $\mathcal{D} = \{(X_i, Y_i, c_i)\}$, frozen backbone $f$, warmup epochs $E_w$, total epochs $E$, refit interval $E_r$, draws per concept $S$
\State Initialize encoder $g_\theta$ and head $h_\phi$
\For{epoch $= 1, \dots, E_w$} \Comment{Warmup}
  \For{batch $(X, Y, c) \in \mathcal{D}$}
    \State $M \gets g_\theta(f(X))$, \quad $\hat{Y} \gets h_\phi(M)$
    \State Update $(\theta, \phi)$ on $\mathcal{L}_{\text{task}} + \lambda_{\text{cls}} \mathcal{L}_{\text{cls}}$
  \EndFor
\EndFor
\State Fit $\mathcal{F}_\psi$ on $\mathcal{L}_{\text{flow}} = -\mathbb{E}\big[\sum_j \log p_\psi(M_j \mid c_j, X)\big]$ with $M$ held fixed \Comment{Intermission}
\For{epoch $= E_w + 1, \dots, E$} \Comment{Alignment}
  \If{$E_r$ epochs have passed since the last fit}
    \State Refit $\mathcal{F}_\psi$ on the current encoder's concept representations
  \EndIf
  \For{batch $(X, Y, c) \in \mathcal{D}$}
    \State $M \gets g_\theta(f(X))$, \quad $\hat{Y} \gets h_\phi(M)$
    \State Draw $S$ counterfactuals $\tilde{M}_j \sim \mathcal{F}_\psi(\cdot \mid 1 - c_j, X)$ for every concept $j$ \Comment{no gradient to $\psi$}
    \State Compute $\Delta^0$, $\Delta^0_j$, $\Delta^{\text{cf}}_j$, and $\mathcal{L}_{\text{align}}$ (Sec.~\ref{sec:method-criteria})
    \State Update $(\theta, \phi)$ on $\mathcal{L}_{\text{wrap}} = \mathcal{L}_{\text{task}} + \lambda_{\text{cls}} \mathcal{L}_{\text{cls}} + \mathcal{L}_{\text{align}}$
  \EndFor
\EndFor
\end{algorithmic}
\end{algorithm}

\subsection{Loss weights}

Loss weights for each domain are shown in Table~\ref{tab:loss-weights}. The task loss has weight 1 everywhere, and the four alignment losses share a common weight. The presence threshold is set as $\beta = 1/\sum_j c_j$ for equal share across each active concept. This gives $\beta = 0.5$ on aircraft control, where two concepts are active per input, and $\beta = 1$ on CarRacing, where one concept is active for any given input.

\begin{table}[H]
\centering
\small
\caption{Loss weights per domain. The CUB-200 models train without causal alignment losses.}
\label{tab:loss-weights}
\begin{tabular}{lcccccc}
\toprule
Domain & $\lambda_{\text{cls}}$ & $\lambda_{\text{nec}}$ & $\lambda_{\text{suf}}$ & $\lambda_{\text{ind}}$ & $\lambda_{\text{pres}}$ & $\beta$ \\
\midrule
Aircraft control & 1.0 & 0.1 & 0.1 & 0.1 & 0.1 & 0.5 \\
CarRacing & 1.0 & 0.1 & 0.1 & 0.1 & 0.1 & 1.0 \\
CUB-200, CCW-Net & 1.0 & --- & --- & --- & --- & --- \\
CUB-200, CBM & 0.01 & --- & --- & --- & --- & --- \\
\bottomrule
\end{tabular}
\end{table}

\subsection{Optimizers and learning rates}

CCW-Net and the mediator model each use Adam with no weight decay, and the mediator model uses the wrapper's batch size (Table~\ref{tab:optimizer-hparams}).

\begin{table}[H]
\centering
\small
\caption{Optimizer settings per domain. The CUB-200 models use no mediator model as they had no causal alignment losses active. App.~\ref{sec:appendix-ablations} lists the batch size of each CarRacing ablation.}
\label{tab:optimizer-hparams}
\begin{tabular}{lccc}
\toprule
Domain & Wrapper learning rate & Flow learning rate & Batch size \\
\midrule
Aircraft control & $5 \times 10^{-4}$ & $1 \times 10^{-3}$ & 2048 \\
CarRacing, Table~\ref{tab:carracing-per-state} & $1 \times 10^{-3}$ & $1 \times 10^{-3}$ & 32 \\
CarRacing, App.~\ref{sec:appendix-ablations} canonical & $1 \times 10^{-3}$ & $1 \times 10^{-3}$ & 128 \\
CUB-200 & $5 \times 10^{-4}$ & --- & 512 \\
\bottomrule
\end{tabular}
\end{table}

\subsection{Training schedule}

\begin{table}[H]
\centering
\small
\caption{Training schedule per domain. The mediator model is fit for 50 epochs at the start of alignment and refit every 10 alignment epochs to refresh with the current concept representations. Each alignment step uses $S = 8$ counterfactual draws per concept. The CUB-200 models train the task and concept losses for all 100 epochs.}
\label{tab:training-schedule}
\begin{tabular}{lcccc}
\toprule
Domain & Warmup epochs $E_w$ & Total epochs $E$ & Flow fit epochs & Refit interval $E_r$ \\
\midrule
Aircraft control & 10 & 100 & 50 & 10 \\
CarRacing & 10 & 50 & 50 & 10 \\
CUB-200 & 100 & 100 & --- & --- \\
\bottomrule
\end{tabular}
\end{table}

\subsection{Data quantities}

\begin{itemize}[leftmargin=1.3em]
\item Aircraft control: 500K samples for the main models. The data-quantity ablations use sets of 5K, 50K, and 5M samples.
\item CarRacing: 15,102 samples from 20 episodes.
\item CUB-200: 11,788 images, split into 72\% train, 13\% validation, and 15\% test (8,516, 1,503, and 1,769 images).
\end{itemize}

\section{Task details}
\label{sec:task-details}

\subsection{Aircraft control}
\label{sec:aircraft-task}

In extended trail formation \citep{aetcman11-251-2024,aetcman11-248-2025}, a chase aircraft maintains a volume of space behind a lead aircraft that flies dynamic maneuvers. The simulator contains a high-fidelity aircraft dynamics model with six degrees of freedom \citep{heidlauf2018verification}. For this experiment, the lead aircraft follows a scripted trajectory, and the chase aircraft is controlled by a reinforcement learning policy trained with proximal policy optimization \citep{schulman2017proximal} using the aircraft environment and codebase of \citet{So-RSS-23}.

\paragraph{Backbone and action space.}
The frozen backbone policy maps the state-based observation $X$ to a two-dimensional action $Y$ of pitch and roll commands.

\paragraph{Concepts.}
$J = 4$ binary concepts drawn from operational pilot vocabulary \citep{aetcman11-251-2024,aetcman11-248-2025} providing real-world external validity to the concept set. No real-world concepts were dropped from the operational ontology for this task:
\begin{itemize}[leftmargin=1.3em]
\item \emph{Lead pursuit}: the chase aircraft's nose points ahead of the lead aircraft.
\item \emph{Lag pursuit}: the chase aircraft's nose points behind the lead aircraft.
\item \emph{Climb}: the aircraft is ascending in altitude.
\item \emph{Dive}: the aircraft is descending in altitude.
\end{itemize}
The pursuit concept set (Lead, Lag) and the vertical concept set (Climb, Dive) are each mutually exclusive. Exactly one concept per set and two concepts overall are active per sample.

\subsection{CarRacing}
\label{sec:carracing-task}

CarRacing-v1 \citep{brockman2016openai} is a top-down driving task with pixel inputs. The agent receives a $96 \times 96$ RGB image of the track and outputs continuous steering, throttle, and brake commands. We use the recorded latents, actions, and concept labels released with CW-Net \citep{kenny2026explainable}.

\paragraph{Backbone and action space.}
The frozen backbone is the convolutional policy from that release, trained end-to-end on CarRacing-v1. The action space is three-dimensional, with steering in $[-1, 1]$, throttle in $[0, 1]$, and brake in $[0, 1]$.

\paragraph{Concepts.}
$J = 3$ binary concepts, which the CW-Net authors labeled manually on over 15,000 examples \citep{kenny2026explainable}:
\begin{itemize}[leftmargin=1.3em]
\item \emph{Straight}: drive straight.
\item \emph{Left}: left turn.
\item \emph{Right}: right turn.
\end{itemize}
The three concepts are mutually exclusive and exhaustive, so exactly one concept is active per sample. The labels do not follow the agent's instantaneous steering command.

\subsection{CUB-200 classification}
\label{sec:cub-task}

CUB-200-2011 \citep{wah2011caltech} is an image classification dataset of 200 bird species. We follow the concept bottleneck setup of \citet{koh2020concept}, in which 112 binary visual attributes, assigned per species, exist between the image and the species label.

\paragraph{Backbone and output space.}
The frozen backbone is Inception-v3, fine-tuned as the joint concept bottleneck model of \citet{koh2020concept} on a species-stratified 80\% of all 11,788 CUB-200 images. This fine-tuning set includes about 80\% of our test images. The wrapper output is a 200-way species prediction.

\paragraph{Concepts.}
$J = 112$ binary visual attributes (e.g., \emph{has yellow belly}, \emph{has pointed beak}, \emph{has streaked wing}). Each species has its own subset of attributes, so the number of active concepts varies by image, with a mean of about 7 over the test split.

\paragraph{Splits.}
We split the images into 72\% train, 13\% validation, and 15\% test (8,516, 1,503, and 1,769 images), with a different split for each seed.

\section{Full ablation matrix}
\label{sec:appendix-ablations}

Tables~\ref{tab:aircraft-full-ablations} and~\ref{tab:carracing-full-ablations} report every ablation on aircraft control and CarRacing. Outside the last group, every CCW-Net model uses an encoder of width 256 (App.~\ref{sec:architecture-details}) and compares against the canonical model in the first row, and the baselines use their own architectures. The last group uses an encoder of width 64, the width of the models in Tables~\ref{tab:aircraft-per-state} and~\ref{tab:carracing-per-state}. All Active and All Inactive average $\rho_{\text{active}}$ and $\rho_{\text{inactive}}$ over the label combinations, weighted by each combination's number of test inputs (App.~\ref{sec:estimator-details}). Loss combinations name the alignment losses used, N (necessity), S (sufficiency), I (independence), and P (presence), and the canonical model uses all four. All values are means over ten seeds with two-sided 95\% Student-$t$ intervals, computed from the trained weights. App.~\ref{sec:estimator-details} gives the estimator's settings.

\begin{landscape}
\subsection{Aircraft control}
\label{sec:aircraft-full-ablations}

{\small
\setlength{\LTcapwidth}{\linewidth}
\begin{longtable}{lrrrrrrrr}
\caption{Aircraft control: full ablation matrix. tMSE is the test action MSE, and concept accuracy is on the Lead/Lag and Climb/Dive concept sets (\%). With a placebo concept added, the placebo carries $36.2 \pm 11.4\%$ of the effect at width 256, and $28.2 \pm 9.8\%$ with and $98.2 \pm 2.9\%$ without the alignment losses at width 64.}
\label{tab:aircraft-full-ablations}\\
\toprule
 & & \multicolumn{2}{c}{Concept acc.\ (\%)} & \multicolumn{2}{c}{Concept Effect (\%)} & \multicolumn{3}{c}{DCI} \\
\cmidrule(lr){3-4}\cmidrule(lr){5-6}\cmidrule(lr){7-9}
Model & tMSE & Lead/Lag & Climb/Dive & All Active $\uparrow$ & All Inactive $\downarrow$ & Disentanglement & Completeness & Informativeness \\
\endfirsthead
\multicolumn{9}{l}{\textit{Table~\ref{tab:aircraft-full-ablations} continued}} \\
\toprule
 & & \multicolumn{2}{c}{Concept acc.\ (\%)} & \multicolumn{2}{c}{Concept Effect (\%)} & \multicolumn{3}{c}{DCI} \\
\cmidrule(lr){3-4}\cmidrule(lr){5-6}\cmidrule(lr){7-9}
Model & tMSE & Lead/Lag & Climb/Dive & All Active $\uparrow$ & All Inactive $\downarrow$ & Disentanglement & Completeness & Informativeness \\
\endhead
\bottomrule
\endlastfoot
\multicolumn{9}{l}{\textit{Ours}} \\
CCW-Net (Ours, 500K samples) & 0.123\,{\scriptsize$\pm$0.001} & 96.4\,{\scriptsize$\pm$0.1} & 99.5\,{\scriptsize$\pm$0.0} & 97.3\,{\scriptsize$\pm$1.1} & 2.7\,{\scriptsize$\pm$1.1} & 0.999\,{\scriptsize$\pm$0.001} & 0.748\,{\scriptsize$\pm$0.050} & 0.979\,{\scriptsize$\pm$0.000} \\
CCW-Net (Ours, 5M samples) & 0.120\,{\scriptsize$\pm$0.000} & 97.5\,{\scriptsize$\pm$0.0} & 99.7\,{\scriptsize$\pm$0.0} & 98.3\,{\scriptsize$\pm$0.9} & 1.7\,{\scriptsize$\pm$0.9} & 1.000\,{\scriptsize$\pm$0.000} & 0.780\,{\scriptsize$\pm$0.043} & 0.986\,{\scriptsize$\pm$0.000} \\
CCW-Net (Ours, scalar $D=1$) & 0.412\,{\scriptsize$\pm$0.003} & 96.2\,{\scriptsize$\pm$0.1} & 99.4\,{\scriptsize$\pm$0.0} & 69.9\,{\scriptsize$\pm$11.2} & 30.1\,{\scriptsize$\pm$11.2} & 0.998\,{\scriptsize$\pm$0.001} & 0.896\,{\scriptsize$\pm$0.065} & 0.978\,{\scriptsize$\pm$0.001} \\
\midrule
\multicolumn{9}{l}{\textit{No causal alignment losses}} \\
CCW-Net ($D=16$) & 0.123\,{\scriptsize$\pm$0.001} & 96.4\,{\scriptsize$\pm$0.1} & 99.5\,{\scriptsize$\pm$0.0} & 89.2\,{\scriptsize$\pm$7.5} & 10.8\,{\scriptsize$\pm$7.5} & 0.999\,{\scriptsize$\pm$0.000} & 0.703\,{\scriptsize$\pm$0.025} & 0.980\,{\scriptsize$\pm$0.000} \\
CCW-Net (scalar $D=1$) & 0.409\,{\scriptsize$\pm$0.002} & 96.1\,{\scriptsize$\pm$0.1} & 99.4\,{\scriptsize$\pm$0.0} & 70.8\,{\scriptsize$\pm$11.9} & 29.2\,{\scriptsize$\pm$11.9} & 0.996\,{\scriptsize$\pm$0.003} & 0.816\,{\scriptsize$\pm$0.089} & 0.978\,{\scriptsize$\pm$0.001} \\
\midrule
\multicolumn{9}{l}{\textit{Architecture baselines}} \\
CW-Net architecture, scalar + alignment & 0.233\,{\scriptsize$\pm$0.031} & 91.0\,{\scriptsize$\pm$0.3} & 98.2\,{\scriptsize$\pm$0.1} & 51.7\,{\scriptsize$\pm$2.3} & 48.3\,{\scriptsize$\pm$2.3} & 0.999\,{\scriptsize$\pm$0.001} & 0.624\,{\scriptsize$\pm$0.078} & 0.929\,{\scriptsize$\pm$0.004} \\
CW-Net (Kenny et al.) & 0.122\,{\scriptsize$\pm$0.002} & 91.2\,{\scriptsize$\pm$0.2} & 98.2\,{\scriptsize$\pm$0.1} & 50.0\,{\scriptsize$\pm$6.2} & 50.0\,{\scriptsize$\pm$6.2} & 0.998\,{\scriptsize$\pm$0.001} & 0.719\,{\scriptsize$\pm$0.064} & 0.947\,{\scriptsize$\pm$0.001} \\
\midrule
\multicolumn{9}{l}{\textit{Disentanglement baselines}} \\
$\beta$-TCVAE ($D=16$) & 0.403\,{\scriptsize$\pm$0.004} & 50.5\,{\scriptsize$\pm$0.8} & 50.0\,{\scriptsize$\pm$1.0} & 46.2\,{\scriptsize$\pm$7.6} & 53.8\,{\scriptsize$\pm$7.6} & 0.072\,{\scriptsize$\pm$0.056} & 0.743\,{\scriptsize$\pm$0.024} & 0.791\,{\scriptsize$\pm$0.013} \\
iVAE ($D=16$) & 0.401\,{\scriptsize$\pm$0.002} & 52.5\,{\scriptsize$\pm$8.8} & 49.1\,{\scriptsize$\pm$7.6} & 47.6\,{\scriptsize$\pm$2.4} & 52.4\,{\scriptsize$\pm$2.4} & 1.000\,{\scriptsize$\pm$0.000} & 0.891\,{\scriptsize$\pm$0.051} & 1.000\,{\scriptsize$\pm$0.000} \\
\midrule
\multicolumn{9}{l}{\textit{Data quantity}} \\
5K samples & 0.378\,{\scriptsize$\pm$0.012} & 91.7\,{\scriptsize$\pm$0.3} & 98.0\,{\scriptsize$\pm$0.4} & 77.2\,{\scriptsize$\pm$8.4} & 22.8\,{\scriptsize$\pm$8.4} & 0.981\,{\scriptsize$\pm$0.010} & 0.718\,{\scriptsize$\pm$0.056} & 0.954\,{\scriptsize$\pm$0.009} \\
50K samples & 0.133\,{\scriptsize$\pm$0.003} & 94.6\,{\scriptsize$\pm$0.1} & 99.0\,{\scriptsize$\pm$0.0} & 92.0\,{\scriptsize$\pm$3.0} & 8.0\,{\scriptsize$\pm$3.0} & 0.992\,{\scriptsize$\pm$0.003} & 0.740\,{\scriptsize$\pm$0.046} & 0.967\,{\scriptsize$\pm$0.002} \\
\midrule
\multicolumn{9}{l}{\textit{Label fraction}} \\
10\% labels & 0.125\,{\scriptsize$\pm$0.001} & 95.2\,{\scriptsize$\pm$0.1} & 99.2\,{\scriptsize$\pm$0.0} & 93.3\,{\scriptsize$\pm$5.1} & 6.7\,{\scriptsize$\pm$5.1} & 0.999\,{\scriptsize$\pm$0.001} & 0.723\,{\scriptsize$\pm$0.031} & 0.972\,{\scriptsize$\pm$0.000} \\
25\% labels & 0.124\,{\scriptsize$\pm$0.001} & 95.8\,{\scriptsize$\pm$0.1} & 99.4\,{\scriptsize$\pm$0.0} & 97.0\,{\scriptsize$\pm$1.4} & 3.0\,{\scriptsize$\pm$1.4} & 0.998\,{\scriptsize$\pm$0.001} & 0.722\,{\scriptsize$\pm$0.056} & 0.976\,{\scriptsize$\pm$0.000} \\
50\% labels & 0.124\,{\scriptsize$\pm$0.001} & 96.1\,{\scriptsize$\pm$0.1} & 99.4\,{\scriptsize$\pm$0.0} & 95.2\,{\scriptsize$\pm$3.3} & 4.8\,{\scriptsize$\pm$3.3} & 0.999\,{\scriptsize$\pm$0.000} & 0.743\,{\scriptsize$\pm$0.052} & 0.978\,{\scriptsize$\pm$0.000} \\
\midrule
\multicolumn{9}{l}{\textit{Label noise}} \\
10\% noise & 0.121\,{\scriptsize$\pm$0.000} & 96.2\,{\scriptsize$\pm$0.1} & 99.4\,{\scriptsize$\pm$0.0} & 71.6\,{\scriptsize$\pm$8.0} & 28.4\,{\scriptsize$\pm$8.0} & 1.000\,{\scriptsize$\pm$0.000} & 0.773\,{\scriptsize$\pm$0.031} & 0.978\,{\scriptsize$\pm$0.000} \\
20\% noise & 0.121\,{\scriptsize$\pm$0.001} & 95.7\,{\scriptsize$\pm$0.1} & 99.3\,{\scriptsize$\pm$0.0} & 66.1\,{\scriptsize$\pm$8.3} & 33.9\,{\scriptsize$\pm$8.3} & 0.999\,{\scriptsize$\pm$0.002} & 0.861\,{\scriptsize$\pm$0.047} & 0.975\,{\scriptsize$\pm$0.000} \\
\midrule
\multicolumn{9}{l}{\textit{Loss combinations (without P)}} \\
N & 0.123\,{\scriptsize$\pm$0.001} & 96.4\,{\scriptsize$\pm$0.1} & 99.5\,{\scriptsize$\pm$0.0} & 91.9\,{\scriptsize$\pm$7.8} & 8.1\,{\scriptsize$\pm$7.8} & 0.998\,{\scriptsize$\pm$0.001} & 0.746\,{\scriptsize$\pm$0.023} & 0.979\,{\scriptsize$\pm$0.000} \\
S & 0.123\,{\scriptsize$\pm$0.001} & 96.4\,{\scriptsize$\pm$0.1} & 99.5\,{\scriptsize$\pm$0.0} & 93.8\,{\scriptsize$\pm$4.3} & 6.2\,{\scriptsize$\pm$4.3} & 0.998\,{\scriptsize$\pm$0.001} & 0.711\,{\scriptsize$\pm$0.035} & 0.979\,{\scriptsize$\pm$0.000} \\
I & 0.123\,{\scriptsize$\pm$0.001} & 96.4\,{\scriptsize$\pm$0.1} & 99.5\,{\scriptsize$\pm$0.0} & 92.3\,{\scriptsize$\pm$5.0} & 7.7\,{\scriptsize$\pm$5.0} & 0.999\,{\scriptsize$\pm$0.001} & 0.684\,{\scriptsize$\pm$0.029} & 0.979\,{\scriptsize$\pm$0.000} \\
N+S & 0.123\,{\scriptsize$\pm$0.001} & 96.4\,{\scriptsize$\pm$0.1} & 99.5\,{\scriptsize$\pm$0.0} & 96.4\,{\scriptsize$\pm$1.3} & 3.6\,{\scriptsize$\pm$1.3} & 0.998\,{\scriptsize$\pm$0.001} & 0.734\,{\scriptsize$\pm$0.043} & 0.980\,{\scriptsize$\pm$0.001} \\
N+I & 0.123\,{\scriptsize$\pm$0.001} & 96.4\,{\scriptsize$\pm$0.1} & 99.5\,{\scriptsize$\pm$0.0} & 94.9\,{\scriptsize$\pm$4.9} & 5.1\,{\scriptsize$\pm$4.9} & 0.998\,{\scriptsize$\pm$0.001} & 0.715\,{\scriptsize$\pm$0.017} & 0.979\,{\scriptsize$\pm$0.000} \\
S+I & 0.123\,{\scriptsize$\pm$0.000} & 96.4\,{\scriptsize$\pm$0.1} & 99.5\,{\scriptsize$\pm$0.0} & 94.7\,{\scriptsize$\pm$2.9} & 5.3\,{\scriptsize$\pm$2.9} & 0.998\,{\scriptsize$\pm$0.001} & 0.703\,{\scriptsize$\pm$0.028} & 0.979\,{\scriptsize$\pm$0.000} \\
N+S+I & 0.123\,{\scriptsize$\pm$0.001} & 96.4\,{\scriptsize$\pm$0.1} & 99.5\,{\scriptsize$\pm$0.0} & 98.0\,{\scriptsize$\pm$0.7} & 2.0\,{\scriptsize$\pm$0.7} & 0.998\,{\scriptsize$\pm$0.001} & 0.714\,{\scriptsize$\pm$0.055} & 0.979\,{\scriptsize$\pm$0.000} \\
\midrule
\multicolumn{9}{l}{\textit{Loss combinations (with P)}} \\
P & 0.124\,{\scriptsize$\pm$0.001} & 96.4\,{\scriptsize$\pm$0.1} & 99.5\,{\scriptsize$\pm$0.0} & 93.0\,{\scriptsize$\pm$3.1} & 7.0\,{\scriptsize$\pm$3.1} & 0.998\,{\scriptsize$\pm$0.001} & 0.708\,{\scriptsize$\pm$0.056} & 0.979\,{\scriptsize$\pm$0.000} \\
N+P & 0.123\,{\scriptsize$\pm$0.001} & 96.4\,{\scriptsize$\pm$0.1} & 99.5\,{\scriptsize$\pm$0.0} & 93.3\,{\scriptsize$\pm$6.0} & 6.7\,{\scriptsize$\pm$6.0} & 0.998\,{\scriptsize$\pm$0.002} & 0.747\,{\scriptsize$\pm$0.033} & 0.980\,{\scriptsize$\pm$0.000} \\
S+P & 0.123\,{\scriptsize$\pm$0.001} & 96.4\,{\scriptsize$\pm$0.1} & 99.5\,{\scriptsize$\pm$0.0} & 93.1\,{\scriptsize$\pm$3.9} & 6.9\,{\scriptsize$\pm$3.9} & 0.998\,{\scriptsize$\pm$0.001} & 0.720\,{\scriptsize$\pm$0.041} & 0.979\,{\scriptsize$\pm$0.000} \\
I+P & 0.124\,{\scriptsize$\pm$0.001} & 96.4\,{\scriptsize$\pm$0.1} & 99.5\,{\scriptsize$\pm$0.0} & 89.5\,{\scriptsize$\pm$4.9} & 10.5\,{\scriptsize$\pm$4.9} & 0.999\,{\scriptsize$\pm$0.001} & 0.699\,{\scriptsize$\pm$0.031} & 0.979\,{\scriptsize$\pm$0.000} \\
N+S+P & 0.124\,{\scriptsize$\pm$0.001} & 96.4\,{\scriptsize$\pm$0.1} & 99.5\,{\scriptsize$\pm$0.0} & 94.8\,{\scriptsize$\pm$3.2} & 5.2\,{\scriptsize$\pm$3.2} & 0.998\,{\scriptsize$\pm$0.001} & 0.750\,{\scriptsize$\pm$0.069} & 0.979\,{\scriptsize$\pm$0.000} \\
N+I+P & 0.124\,{\scriptsize$\pm$0.001} & 96.4\,{\scriptsize$\pm$0.1} & 99.5\,{\scriptsize$\pm$0.0} & 94.4\,{\scriptsize$\pm$5.5} & 5.6\,{\scriptsize$\pm$5.5} & 0.999\,{\scriptsize$\pm$0.001} & 0.757\,{\scriptsize$\pm$0.049} & 0.979\,{\scriptsize$\pm$0.000} \\
\midrule
\multicolumn{9}{l}{\textit{Concept ontology, merged concepts}} \\
Merge Climb+Dive & 0.122\,{\scriptsize$\pm$0.000} & 96.4\,{\scriptsize$\pm$0.1} & 55.6\,{\scriptsize$\pm$0.0} & 97.4\,{\scriptsize$\pm$1.1} & 2.6\,{\scriptsize$\pm$1.1} & 0.823\,{\scriptsize$\pm$0.095} & 0.573\,{\scriptsize$\pm$0.047} & 0.842\,{\scriptsize$\pm$0.007} \\
Merge Lead+Lag & 0.121\,{\scriptsize$\pm$0.001} & 41.4\,{\scriptsize$\pm$0.0} & 99.5\,{\scriptsize$\pm$0.0} & 95.6\,{\scriptsize$\pm$5.1} & 4.4\,{\scriptsize$\pm$5.1} & 0.885\,{\scriptsize$\pm$0.067} & 0.566\,{\scriptsize$\pm$0.041} & 0.833\,{\scriptsize$\pm$0.005} \\
\midrule
\multicolumn{9}{l}{\textit{Concept ontology, removed concepts}} \\
Remove Climb & 0.124\,{\scriptsize$\pm$0.001} & 96.4\,{\scriptsize$\pm$0.1} & 44.4\,{\scriptsize$\pm$0.0} & 93.9\,{\scriptsize$\pm$1.5} & 6.1\,{\scriptsize$\pm$1.5} & 0.998\,{\scriptsize$\pm$0.001} & 0.764\,{\scriptsize$\pm$0.042} & 0.979\,{\scriptsize$\pm$0.000} \\
Remove Dive & 0.124\,{\scriptsize$\pm$0.001} & 96.4\,{\scriptsize$\pm$0.1} & 55.6\,{\scriptsize$\pm$0.0} & 94.6\,{\scriptsize$\pm$3.8} & 5.4\,{\scriptsize$\pm$3.8} & 0.999\,{\scriptsize$\pm$0.001} & 0.792\,{\scriptsize$\pm$0.042} & 0.980\,{\scriptsize$\pm$0.000} \\
Remove Lag & 0.123\,{\scriptsize$\pm$0.001} & 41.4\,{\scriptsize$\pm$0.0} & 99.4\,{\scriptsize$\pm$0.0} & 79.9\,{\scriptsize$\pm$11.4} & 20.1\,{\scriptsize$\pm$11.4} & 0.999\,{\scriptsize$\pm$0.001} & 0.750\,{\scriptsize$\pm$0.039} & 0.980\,{\scriptsize$\pm$0.000} \\
Remove Lead & 0.123\,{\scriptsize$\pm$0.001} & 58.6\,{\scriptsize$\pm$0.0} & 99.5\,{\scriptsize$\pm$0.0} & 85.9\,{\scriptsize$\pm$12.0} & 14.1\,{\scriptsize$\pm$12.0} & 0.998\,{\scriptsize$\pm$0.001} & 0.774\,{\scriptsize$\pm$0.049} & 0.980\,{\scriptsize$\pm$0.000} \\
\midrule
\multicolumn{9}{l}{\textit{Placebo}} \\
Placebo concept added & 0.123\,{\scriptsize$\pm$0.001} & 96.4\,{\scriptsize$\pm$0.1} & 99.5\,{\scriptsize$\pm$0.0} & 61.1\,{\scriptsize$\pm$11.4} & 38.9\,{\scriptsize$\pm$11.4} & 0.999\,{\scriptsize$\pm$0.001} & 0.770\,{\scriptsize$\pm$0.028} & 0.980\,{\scriptsize$\pm$0.000} \\
\midrule
\multicolumn{9}{l}{\textit{Encoder width 64}} \\
CCW-Net (Ours), $D=16$ & 0.125\,{\scriptsize$\pm$0.001} & 95.7\,{\scriptsize$\pm$0.1} & 99.3\,{\scriptsize$\pm$0.0} & 94.5\,{\scriptsize$\pm$2.2} & 5.5\,{\scriptsize$\pm$2.2} & 0.999\,{\scriptsize$\pm$0.001} & 0.719\,{\scriptsize$\pm$0.041} & 0.975\,{\scriptsize$\pm$0.001} \\
CCW-Net (Ours), $D=8$ & 0.125\,{\scriptsize$\pm$0.001} & 95.6\,{\scriptsize$\pm$0.1} & 99.3\,{\scriptsize$\pm$0.0} & 93.1\,{\scriptsize$\pm$5.2} & 6.9\,{\scriptsize$\pm$5.2} & 0.999\,{\scriptsize$\pm$0.000} & 0.717\,{\scriptsize$\pm$0.053} & 0.974\,{\scriptsize$\pm$0.001} \\
10\% label noise & 0.121\,{\scriptsize$\pm$0.001} & 95.3\,{\scriptsize$\pm$0.1} & 99.2\,{\scriptsize$\pm$0.0} & 70.4\,{\scriptsize$\pm$5.5} & 29.6\,{\scriptsize$\pm$5.5} & 1.000\,{\scriptsize$\pm$0.000} & 0.762\,{\scriptsize$\pm$0.031} & 0.973\,{\scriptsize$\pm$0.000} \\
10\% label noise, no alignment & 0.121\,{\scriptsize$\pm$0.000} & 95.3\,{\scriptsize$\pm$0.1} & 99.2\,{\scriptsize$\pm$0.0} & 59.6\,{\scriptsize$\pm$12.2} & 40.4\,{\scriptsize$\pm$12.2} & 1.000\,{\scriptsize$\pm$0.000} & 0.750\,{\scriptsize$\pm$0.040} & 0.973\,{\scriptsize$\pm$0.001} \\
Placebo concept added & 0.124\,{\scriptsize$\pm$0.001} & 95.6\,{\scriptsize$\pm$0.1} & 99.3\,{\scriptsize$\pm$0.0} & 67.0\,{\scriptsize$\pm$10.8} & 33.0\,{\scriptsize$\pm$10.8} & 0.999\,{\scriptsize$\pm$0.001} & 0.711\,{\scriptsize$\pm$0.043} & 0.975\,{\scriptsize$\pm$0.001} \\
Placebo concept added, no alignment & 0.121\,{\scriptsize$\pm$0.001} & 95.7\,{\scriptsize$\pm$0.1} & 99.4\,{\scriptsize$\pm$0.0} & 0.9\,{\scriptsize$\pm$1.8} & 99.1\,{\scriptsize$\pm$1.8} & 0.999\,{\scriptsize$\pm$0.001} & 0.685\,{\scriptsize$\pm$0.020} & 0.975\,{\scriptsize$\pm$0.001} \\
\end{longtable}}
\end{landscape}

\begin{landscape}
\subsection{CarRacing}
\label{sec:carracing-full-ablations}

{\small
\setlength{\LTcapwidth}{\linewidth}
\begin{longtable}{lrrrrrrr}
\caption{CarRacing: full ablation matrix. tMSE is the test action MSE, and concept accuracy is on the Straight/Left/Right concept set (\%). Every group except the last uses batch size 128 for the canonical, $D=16$ no-alignment, loss-combination, label-noise, and placebo rows and 32 for the other rows. The last group uses batch size 32. Every row with the presence loss uses $\beta=1$ except CCW-Net (Ours, scalar $D=1$) and the two CW-Net rows with alignment, which use $\beta=0.5$. For the Remove rows, the averages cover the label combinations with an active concept, since the removed concept's own combination has none. \S\ In the Merge rows, the estimator scores only the merged concept's label combination, because the remaining combination has no comparison that holds the merged concept fixed. With a placebo concept added, the placebo carries $34.6 \pm 6.8\%$ of the effect.}
\label{tab:carracing-full-ablations}\\
\toprule
 & & & \multicolumn{2}{c}{Concept Effect (\%)} & \multicolumn{3}{c}{DCI} \\
\cmidrule(lr){4-5}\cmidrule(lr){6-8}
Model & tMSE & Concept acc.\ (\%) & All Active $\uparrow$ & All Inactive $\downarrow$ & Disentanglement & Completeness & Informativeness \\
\endfirsthead
\multicolumn{8}{l}{\textit{Table~\ref{tab:carracing-full-ablations} continued}} \\
\toprule
 & & & \multicolumn{2}{c}{Concept Effect (\%)} & \multicolumn{3}{c}{DCI} \\
\cmidrule(lr){4-5}\cmidrule(lr){6-8}
Model & tMSE & Concept acc.\ (\%) & All Active $\uparrow$ & All Inactive $\downarrow$ & Disentanglement & Completeness & Informativeness \\
\endhead
\bottomrule
\endlastfoot
\multicolumn{8}{l}{\textit{Ours}} \\
CCW-Net (Ours, $\beta=1.0$) & 0.0025\,{\scriptsize$\pm$0.0004} & 93.4\,{\scriptsize$\pm$0.2} & 75.3\,{\scriptsize$\pm$3.8} & 24.7\,{\scriptsize$\pm$3.8} & 0.914\,{\scriptsize$\pm$0.015} & 0.706\,{\scriptsize$\pm$0.042} & 0.950\,{\scriptsize$\pm$0.004} \\
CCW-Net (Ours, scalar $D=1$) & 0.0420\,{\scriptsize$\pm$0.0006} & 93.0\,{\scriptsize$\pm$0.3} & 92.4\,{\scriptsize$\pm$2.7} & 7.6\,{\scriptsize$\pm$2.7} & 0.866\,{\scriptsize$\pm$0.021} & 0.866\,{\scriptsize$\pm$0.021} & 0.947\,{\scriptsize$\pm$0.005} \\
\midrule
\multicolumn{8}{l}{\textit{No causal alignment losses}} \\
CCW-Net ($D=16$) & 0.0032\,{\scriptsize$\pm$0.0004} & 93.4\,{\scriptsize$\pm$0.3} & 63.5\,{\scriptsize$\pm$9.0} & 36.5\,{\scriptsize$\pm$9.0} & 0.913\,{\scriptsize$\pm$0.011} & 0.680\,{\scriptsize$\pm$0.034} & 0.950\,{\scriptsize$\pm$0.003} \\
CCW-Net (scalar $D=1$) & 0.0419\,{\scriptsize$\pm$0.0006} & 93.0\,{\scriptsize$\pm$0.4} & 90.0\,{\scriptsize$\pm$5.5} & 10.0\,{\scriptsize$\pm$5.5} & 0.870\,{\scriptsize$\pm$0.012} & 0.871\,{\scriptsize$\pm$0.012} & 0.948\,{\scriptsize$\pm$0.003} \\
\midrule
\multicolumn{8}{l}{\textit{Architecture baselines}} \\
CW-Net architecture, scalar + alignment & 0.0276\,{\scriptsize$\pm$0.0089} & 89.8\,{\scriptsize$\pm$1.3} & 37.2\,{\scriptsize$\pm$6.7} & 62.8\,{\scriptsize$\pm$6.7} & 0.085\,{\scriptsize$\pm$0.061} & 0.560\,{\scriptsize$\pm$0.083} & 0.928\,{\scriptsize$\pm$0.009} \\
CW-Net (Kenny et al.) & 0.0015\,{\scriptsize$\pm$0.0007} & 90.0\,{\scriptsize$\pm$1.3} & 33.1\,{\scriptsize$\pm$1.6} & 66.9\,{\scriptsize$\pm$1.6} & 0.078\,{\scriptsize$\pm$0.063} & 0.626\,{\scriptsize$\pm$0.082} & 0.928\,{\scriptsize$\pm$0.008} \\
CW-Net architecture, $D=16$ + alignment & 0.0058\,{\scriptsize$\pm$0.0015} & 92.3\,{\scriptsize$\pm$0.5} & 52.2\,{\scriptsize$\pm$2.9} & 47.8\,{\scriptsize$\pm$2.9} & 0.763\,{\scriptsize$\pm$0.024} & 0.710\,{\scriptsize$\pm$0.021} & 0.938\,{\scriptsize$\pm$0.005} \\
CW-Net architecture, $D=16$ & 0.0004\,{\scriptsize$\pm$0.0001} & 92.4\,{\scriptsize$\pm$0.4} & 43.3\,{\scriptsize$\pm$2.4} & 56.7\,{\scriptsize$\pm$2.4} & 0.772\,{\scriptsize$\pm$0.031} & 0.678\,{\scriptsize$\pm$0.034} & 0.938\,{\scriptsize$\pm$0.003} \\
\midrule
\multicolumn{8}{l}{\textit{Disentanglement baselines}} \\
$\beta$-TCVAE ($D=1$) & 0.0290\,{\scriptsize$\pm$0.0024} & 31.0\,{\scriptsize$\pm$7.6} & 37.9\,{\scriptsize$\pm$8.1} & 62.1\,{\scriptsize$\pm$8.1} & 0.026\,{\scriptsize$\pm$0.019} & 0.472\,{\scriptsize$\pm$0.132} & 0.890\,{\scriptsize$\pm$0.014} \\
$\beta$-TCVAE ($D=16$) & 0.0293\,{\scriptsize$\pm$0.0006} & 32.2\,{\scriptsize$\pm$1.8} & 51.7\,{\scriptsize$\pm$8.8} & 48.3\,{\scriptsize$\pm$8.8} & 0.047\,{\scriptsize$\pm$0.012} & 0.813\,{\scriptsize$\pm$0.025} & 0.894\,{\scriptsize$\pm$0.005} \\
iVAE ($D=1$) & 0.0277\,{\scriptsize$\pm$0.0009} & 32.4\,{\scriptsize$\pm$13.0} & 39.0\,{\scriptsize$\pm$7.0} & 61.0\,{\scriptsize$\pm$7.0} & 0.022\,{\scriptsize$\pm$0.018} & 0.354\,{\scriptsize$\pm$0.079} & 0.897\,{\scriptsize$\pm$0.009} \\
iVAE ($D=16$) & 0.0191\,{\scriptsize$\pm$0.0013} & 45.1\,{\scriptsize$\pm$11.0} & 37.3\,{\scriptsize$\pm$4.8} & 62.7\,{\scriptsize$\pm$4.8} & 0.250\,{\scriptsize$\pm$0.066} & 0.689\,{\scriptsize$\pm$0.019} & 0.911\,{\scriptsize$\pm$0.006} \\
\midrule
\multicolumn{8}{l}{\textit{Label fraction}} \\
10\% labels & 0.0026\,{\scriptsize$\pm$0.0003} & 90.2\,{\scriptsize$\pm$0.5} & 77.9\,{\scriptsize$\pm$4.9} & 22.1\,{\scriptsize$\pm$4.9} & 0.909\,{\scriptsize$\pm$0.011} & 0.675\,{\scriptsize$\pm$0.030} & 0.930\,{\scriptsize$\pm$0.007} \\
25\% labels & 0.0030\,{\scriptsize$\pm$0.0005} & 91.2\,{\scriptsize$\pm$0.4} & 75.6\,{\scriptsize$\pm$7.3} & 24.4\,{\scriptsize$\pm$7.3} & 0.910\,{\scriptsize$\pm$0.012} & 0.684\,{\scriptsize$\pm$0.046} & 0.936\,{\scriptsize$\pm$0.007} \\
50\% labels & 0.0032\,{\scriptsize$\pm$0.0003} & 92.0\,{\scriptsize$\pm$0.4} & 77.1\,{\scriptsize$\pm$6.9} & 22.9\,{\scriptsize$\pm$6.9} & 0.926\,{\scriptsize$\pm$0.023} & 0.681\,{\scriptsize$\pm$0.050} & 0.941\,{\scriptsize$\pm$0.003} \\
\midrule
\multicolumn{8}{l}{\textit{Label noise}} \\
10\% noise & 0.0024\,{\scriptsize$\pm$0.0002} & 92.3\,{\scriptsize$\pm$0.3} & 61.8\,{\scriptsize$\pm$9.3} & 38.2\,{\scriptsize$\pm$9.3} & 0.923\,{\scriptsize$\pm$0.009} & 0.753\,{\scriptsize$\pm$0.049} & 0.944\,{\scriptsize$\pm$0.004} \\
20\% noise & 0.0023\,{\scriptsize$\pm$0.0001} & 91.7\,{\scriptsize$\pm$0.4} & 44.9\,{\scriptsize$\pm$8.3} & 55.1\,{\scriptsize$\pm$8.3} & 0.885\,{\scriptsize$\pm$0.020} & 0.719\,{\scriptsize$\pm$0.044} & 0.940\,{\scriptsize$\pm$0.005} \\
\midrule
\multicolumn{8}{l}{\textit{Loss combinations (without P)}} \\
N & 0.0031\,{\scriptsize$\pm$0.0004} & 93.2\,{\scriptsize$\pm$0.2} & 77.2\,{\scriptsize$\pm$6.1} & 22.8\,{\scriptsize$\pm$6.1} & 0.923\,{\scriptsize$\pm$0.012} & 0.685\,{\scriptsize$\pm$0.034} & 0.948\,{\scriptsize$\pm$0.004} \\
S & 0.0025\,{\scriptsize$\pm$0.0004} & 93.4\,{\scriptsize$\pm$0.2} & 72.4\,{\scriptsize$\pm$7.0} & 27.6\,{\scriptsize$\pm$7.0} & 0.924\,{\scriptsize$\pm$0.010} & 0.676\,{\scriptsize$\pm$0.024} & 0.949\,{\scriptsize$\pm$0.003} \\
I & 0.0031\,{\scriptsize$\pm$0.0002} & 93.4\,{\scriptsize$\pm$0.2} & 63.9\,{\scriptsize$\pm$8.3} & 36.1\,{\scriptsize$\pm$8.3} & 0.919\,{\scriptsize$\pm$0.014} & 0.670\,{\scriptsize$\pm$0.029} & 0.950\,{\scriptsize$\pm$0.004} \\
N+S & 0.0025\,{\scriptsize$\pm$0.0004} & 93.4\,{\scriptsize$\pm$0.3} & 78.1\,{\scriptsize$\pm$4.9} & 21.9\,{\scriptsize$\pm$4.9} & 0.921\,{\scriptsize$\pm$0.014} & 0.725\,{\scriptsize$\pm$0.030} & 0.948\,{\scriptsize$\pm$0.004} \\
N+I & 0.0032\,{\scriptsize$\pm$0.0004} & 93.4\,{\scriptsize$\pm$0.3} & 76.1\,{\scriptsize$\pm$2.2} & 23.9\,{\scriptsize$\pm$2.2} & 0.917\,{\scriptsize$\pm$0.019} & 0.674\,{\scriptsize$\pm$0.041} & 0.948\,{\scriptsize$\pm$0.005} \\
S+I & 0.0026\,{\scriptsize$\pm$0.0005} & 93.2\,{\scriptsize$\pm$0.3} & 72.4\,{\scriptsize$\pm$8.4} & 27.6\,{\scriptsize$\pm$8.4} & 0.927\,{\scriptsize$\pm$0.009} & 0.695\,{\scriptsize$\pm$0.033} & 0.948\,{\scriptsize$\pm$0.002} \\
N+S+I & 0.0026\,{\scriptsize$\pm$0.0002} & 93.3\,{\scriptsize$\pm$0.4} & 74.2\,{\scriptsize$\pm$4.1} & 25.8\,{\scriptsize$\pm$4.1} & 0.926\,{\scriptsize$\pm$0.011} & 0.708\,{\scriptsize$\pm$0.023} & 0.950\,{\scriptsize$\pm$0.003} \\
\midrule
\multicolumn{8}{l}{\textit{Loss combinations (with P)}} \\
P & 0.0033\,{\scriptsize$\pm$0.0005} & 93.3\,{\scriptsize$\pm$0.1} & 71.9\,{\scriptsize$\pm$8.9} & 28.1\,{\scriptsize$\pm$8.9} & 0.919\,{\scriptsize$\pm$0.012} & 0.684\,{\scriptsize$\pm$0.029} & 0.950\,{\scriptsize$\pm$0.005} \\
N+P & 0.0030\,{\scriptsize$\pm$0.0004} & 93.2\,{\scriptsize$\pm$0.2} & 77.4\,{\scriptsize$\pm$4.1} & 22.6\,{\scriptsize$\pm$4.1} & 0.918\,{\scriptsize$\pm$0.012} & 0.682\,{\scriptsize$\pm$0.030} & 0.949\,{\scriptsize$\pm$0.004} \\
S+P & 0.0026\,{\scriptsize$\pm$0.0004} & 93.5\,{\scriptsize$\pm$0.3} & 75.2\,{\scriptsize$\pm$5.9} & 24.8\,{\scriptsize$\pm$5.9} & 0.922\,{\scriptsize$\pm$0.016} & 0.689\,{\scriptsize$\pm$0.034} & 0.950\,{\scriptsize$\pm$0.004} \\
I+P & 0.0031\,{\scriptsize$\pm$0.0004} & 93.5\,{\scriptsize$\pm$0.3} & 67.6\,{\scriptsize$\pm$9.6} & 32.4\,{\scriptsize$\pm$9.6} & 0.928\,{\scriptsize$\pm$0.009} & 0.726\,{\scriptsize$\pm$0.037} & 0.951\,{\scriptsize$\pm$0.004} \\
N+S+P & 0.0026\,{\scriptsize$\pm$0.0004} & 93.3\,{\scriptsize$\pm$0.2} & 77.7\,{\scriptsize$\pm$5.6} & 22.3\,{\scriptsize$\pm$5.6} & 0.922\,{\scriptsize$\pm$0.015} & 0.698\,{\scriptsize$\pm$0.038} & 0.948\,{\scriptsize$\pm$0.002} \\
N+I+P & 0.0031\,{\scriptsize$\pm$0.0004} & 93.3\,{\scriptsize$\pm$0.3} & 76.0\,{\scriptsize$\pm$4.7} & 24.0\,{\scriptsize$\pm$4.7} & 0.923\,{\scriptsize$\pm$0.011} & 0.699\,{\scriptsize$\pm$0.032} & 0.949\,{\scriptsize$\pm$0.003} \\
S+I+P & 0.0025\,{\scriptsize$\pm$0.0004} & 93.3\,{\scriptsize$\pm$0.2} & 76.2\,{\scriptsize$\pm$3.5} & 23.8\,{\scriptsize$\pm$3.5} & 0.927\,{\scriptsize$\pm$0.017} & 0.668\,{\scriptsize$\pm$0.043} & 0.950\,{\scriptsize$\pm$0.002} \\
\midrule
\multicolumn{8}{l}{\textit{Concept ontology, merged concepts}} \\
Merge Left+Right\,\S & 0.0023\,{\scriptsize$\pm$0.0002} & 84.4\,{\scriptsize$\pm$0.6} & 56.9\,{\scriptsize$\pm$9.9} & 43.1\,{\scriptsize$\pm$9.9} & 0.625\,{\scriptsize$\pm$0.097} & 0.553\,{\scriptsize$\pm$0.047} & 0.920\,{\scriptsize$\pm$0.006} \\
Merge Straight+Left\,\S & 0.0015\,{\scriptsize$\pm$0.0003} & 72.1\,{\scriptsize$\pm$0.2} & 95.7\,{\scriptsize$\pm$2.5} & 4.3\,{\scriptsize$\pm$2.5} & 0.629\,{\scriptsize$\pm$0.045} & 0.481\,{\scriptsize$\pm$0.027} & 0.842\,{\scriptsize$\pm$0.011} \\
Merge Straight+Right\,\S & 0.0019\,{\scriptsize$\pm$0.0002} & 84.8\,{\scriptsize$\pm$0.3} & 90.1\,{\scriptsize$\pm$7.2} & 9.9\,{\scriptsize$\pm$7.2} & 0.538\,{\scriptsize$\pm$0.073} & 0.542\,{\scriptsize$\pm$0.039} & 0.909\,{\scriptsize$\pm$0.005} \\
\midrule
\multicolumn{8}{l}{\textit{Concept ontology, removed concepts}} \\
Remove Left & 0.0036\,{\scriptsize$\pm$0.0003} & 72.4\,{\scriptsize$\pm$0.3} & 71.1\,{\scriptsize$\pm$9.1} & 28.9\,{\scriptsize$\pm$9.1} & 0.796\,{\scriptsize$\pm$0.039} & 0.668\,{\scriptsize$\pm$0.039} & 0.945\,{\scriptsize$\pm$0.004} \\
Remove Right & 0.0049\,{\scriptsize$\pm$0.0008} & 85.1\,{\scriptsize$\pm$0.3} & 82.1\,{\scriptsize$\pm$5.4} & 17.9\,{\scriptsize$\pm$5.4} & 0.826\,{\scriptsize$\pm$0.040} & 0.652\,{\scriptsize$\pm$0.027} & 0.944\,{\scriptsize$\pm$0.006} \\
Remove Straight & 0.0085\,{\scriptsize$\pm$0.0013} & 35.0\,{\scriptsize$\pm$0.1} & 65.8\,{\scriptsize$\pm$10.3} & 34.2\,{\scriptsize$\pm$10.3} & 0.637\,{\scriptsize$\pm$0.114} & 0.688\,{\scriptsize$\pm$0.023} & 0.948\,{\scriptsize$\pm$0.004} \\
\midrule
\multicolumn{8}{l}{\textit{Placebo}} \\
Placebo concept added & 0.0055\,{\scriptsize$\pm$0.0008} & 93.2\,{\scriptsize$\pm$0.2} & 51.3\,{\scriptsize$\pm$7.9} & 48.7\,{\scriptsize$\pm$7.9} & 0.935\,{\scriptsize$\pm$0.012} & 0.701\,{\scriptsize$\pm$0.034} & 0.949\,{\scriptsize$\pm$0.004} \\
\midrule
\multicolumn{8}{l}{\textit{Encoder width 64}} \\
CCW-Net (Ours), $D=16$ & 0.0031\,{\scriptsize$\pm$0.0002} & 92.6\,{\scriptsize$\pm$0.5} & 71.9\,{\scriptsize$\pm$5.7} & 28.1\,{\scriptsize$\pm$5.7} & 0.921\,{\scriptsize$\pm$0.011} & 0.686\,{\scriptsize$\pm$0.031} & 0.945\,{\scriptsize$\pm$0.005} \\
CCW-Net (Ours), $D=8$ & 0.0037\,{\scriptsize$\pm$0.0004} & 92.7\,{\scriptsize$\pm$0.4} & 67.8\,{\scriptsize$\pm$6.1} & 32.2\,{\scriptsize$\pm$6.1} & 0.880\,{\scriptsize$\pm$0.010} & 0.673\,{\scriptsize$\pm$0.039} & 0.946\,{\scriptsize$\pm$0.004} \\
\end{longtable}}
\end{landscape}

\end{document}